\documentclass[10pt,twocolumn,letterpaper]{article}

\usepackage{abstract}
\usepackage{iccv}
\usepackage{times}
\usepackage{epsfig}
\usepackage{graphicx}
\usepackage{amsmath}
\usepackage{amssymb}

\usepackage{graphicx}
\usepackage{amsmath}
\usepackage{amssymb}
\usepackage{booktabs}
\usepackage{subfig}

\usepackage{multirow}
\usepackage{multicol}
\usepackage{makecell}

\usepackage[pagebackref=true,breaklinks=true,letterpaper=true,colorlinks,bookmarks=false]{hyperref}

\newcommand{\et}[2]{${#1}^{\pm{#2}}$}

\iccvfinalcopy 

\begin{document}

\title{TM2D: Bimodality Driven 3D Dance Generation via Music-Text Integration}

\author{
{Kehong~Gong\textsuperscript{1}\footnotemark[1]} \qquad \quad 
{Dongze~Lian\textsuperscript{1}\thanks{Equal contribution: {\tt gongkehong@u.nus.edu, dongze@nus.edu.sg }}}  \quad  \qquad {Heng~Chang\textsuperscript{2}}  \quad  \qquad 
{Chuan~Guo\textsuperscript{2}} \quad  \qquad \\
\quad  
{Zihang~Jiang\textsuperscript{1}} \qquad \quad \quad 
{Xinxin~Zuo\textsuperscript{2}}  \qquad  \quad  
{Michael~Bi~Mi\textsuperscript{2}}  \qquad    
{Xinchao~Wang\textsuperscript{1}\footnotemark[2]\thanks{Corresponding author: {\tt  xinchao@nus.edu.sg}}} \quad  \qquad \\ 
\normalsize{\textsuperscript{1} National University of Singapore} \quad \quad
\normalsize{\textsuperscript{2} Huawei Technologies Co., Ltd.}\\
}


\twocolumn[{%
\renewcommand\twocolumn[1][]{#1}%
\maketitle
\ificcvfinal\thispagestyle{empty}\fi
\begin{center}
    \centering
    \small{
     \includegraphics[width=\linewidth]{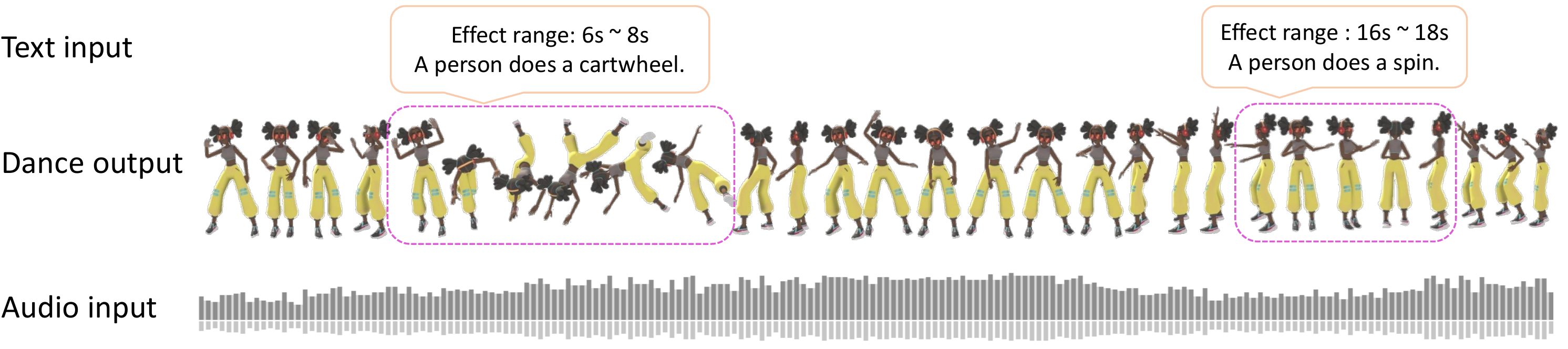}
    \captionof{figure}{Generated 3D dance examples conditioned music and text with our method. Given an audio input and text input with a specific starting time and duration, our method is able to generate a sequence of dance motions that fit the music and text instruction. The character is from Mixamo~\cite{mixamo}.}
    \label{fig: teaser}
    }
\end{center}%
}]
\saythanks

\begin{abstract}
We propose a novel task for generating 3D dance movements that simultaneously incorporate both text and music modalities. Unlike existing works that generate dance movements using a single modality such as music, our goal is to produce richer dance movements guided by the instructive information provided by the text. 
However, the lack of paired motion data with both music and text modalities limits the ability to generate dance movements that integrate both. To alleviate this challenge, we propose to utilize a 3D human motion VQ-VAE to project the motions of the two datasets into a latent space consisting of quantized vectors, which effectively mix the motion tokens from the two datasets with different distributions for training. 
Additionally, we propose a cross-modal transformer to integrate text instructions into motion generation architecture for generating 3D dance movements without degrading the performance of music-conditioned dance generation. 
To better evaluate the quality of the generated motion, we introduce two novel metrics, namely Motion Prediction Distance (MPD) and Freezing Score (FS), to measure the coherence and freezing percentage of the generated motion.
Extensive experiments show that our approach can generate realistic and coherent dance movements conditioned on both text and music while maintaining comparable performance with the two single modalities.
Code is available at \url{https://garfield-kh.github.io/TM2D/}.
\end{abstract}
\section{Introduction}\label{sec: intro}

The music-conditioned dance generation has become a topic of great interest in recent years. The ability to generate dance movements that are synchronized with music has numerous applications, such as behavior understanding, simulation, and benefiting the community of dancers and musicians \cite{lee2019dancing, chen2021choreomaster, li2021ai, siyao2022bailando}. Although music has been used as a guidance to generate dance movements, another important modality cue, text (or language), which provides richer actions and more flexible motion guidance, and is studied in other tasks such as image classification \cite{radford2021learning}, detection \cite{gu2021open}, segmentation \cite{xu2022groupvit}, and text-driven image generation \cite{rombach2022high}, has not been fully explored in dance generation. To this end, we first propose a novel task for generating 3D dance movements that simultaneously incorporate both text and music modalities, enabling the generated human to perform rich dancing movements in accordance with the music and text.

Designing a system pipeline for this bimodality driven 3D dance generation task is non-trivial. There exist two significant challenges to be considered: 
i) the existing datasets only cater to either music-driven (music2dance) \cite{tang2018dance, alemi2017groovenet, zhuang2022music2dance, li2021ai} or text-driven (text2motion) \cite{plappert2016kit, guo2022generating} human motion generation, and no paired 3D dance generation dataset exists that takes into account both music and text. While building a new large-scale paired 3D dance dataset based on music and text is possible, it is time-consuming with fully annotated 3D human motion \cite{li2021ai};
ii) the integration of text into music-conditioned dance generation requires a suitable architecture. However, existing methods that use music as a driving force to generate dance movements might result in temporal-freezing frames or fail to generalize to in-the-wild scenarios \cite{li2021ai, siyao2022bailando}. Therefore, simple integration of text into the existing music-conditioned architecture might pose a risk of degraded dance generation quality in our new task.

To address the first challenge, we take advantage of existing music-dance and text-motion datasets for this new task. However, directly mixing motions from these two datasets would result in inferior performance since the motions from these two datasets are in completely different motion spaces. To overcome this, we propose to utilize a VQ-VAE architecture to project the motions into a consistent and shared latent space. In particular, we build up a shared codebook for all the motions from the training set, and motions from both datasets are now represented as discrete tokens that are implicitly constrained to fall into a shared latent space.
For the second challenge, we propose to utilize a cross-modal transformer architecture that formulates both music2dance and text2motion as sequence-to-sequence translation tasks. This architecture directly translates audio and text features into motion tokens and enables bimodality driven ability by introducing a fusion strategy in the latent space with a shared motion decoder for both tasks. With the shared decoder, audio and text information can be efficiently fused during inference. Our entire cross-modal transformer architecture is both effective and efficient, allowing for the integration of text instructions to generate coherent 3D dance motions, as illustrated in Figure \ref{fig: teaser}.

To better evaluate the coherence of generated dance in our task, we propose a new evaluation metric, Motion Prediction Distance (MPD), which measures the distance between the predicted motion and the ground truth at the time of integrating text, thereby providing a more accurate evaluation of the coherence of frames. Additionally, we introduce a Freezing Score (FS) that quantifies the percentage of temporal freezing frames in dance generation, which is a common problem in music-conditioned dance generation.
To better evaluate the performance of our method in real-world scenarios, we also collect some in-the-wild music data for evaluation. Our method successfully performs dance generation based on both text and music while maintaining comparable performance on the single modality tasks (music2dance, text2motion) compared to other state-of-the-art methods.

In summary, our contributions are as follows:
i) We propose an interesting task of utilizing both music and text for 3D dance generation and propose a pipeline named TM2D (Text-Music to Dance) for this task.
ii) Rather than collecting a new training set, we effectively combine the existing music2dance and text2motion datasets and employ a VQ-VAE framework to encode motions from all training sets to a shared feature space.
iii) We propose a cross-modal transformer as well as a bimodal feature fusion strategy to encode both audio and text features, which is both effective and efficient.
iv) We propose two new metrics, MPD and FS, which efficiently reflect the quality of generated motion.
v) We successfully generate realistic and coherent dance based on both music and text instructions while maintaining comparable performance on the single modality tasks (music2dance, text2motion).

\section{Related Work}\label{sec: related_work}

\subsection{Music to Dance}
Music2Dance is typically divided into 2D and 3D dance generation and has been explored for many years. Recent methods model 3D dance generation from the perspective of network architecture. For instance, Lee \emph{et al.} \cite{lee2018listen} explore the convolutional neural network in casual dilated setting. Lee \emph{et al.} \cite{lee2019dancing} propose a dance unit with Variational Auto-Encoder (VAE). Ren \emph{et al.} \cite{ren2020self} employ recurrent neural network (RNN) and the graph convolution network is introduced in \cite{ferreira2021learning}. As for the 3D dance generation, \cite{ye2020choreonet, ahn2020generative, kritsis2022danceconv} implement convolutional neural network, \cite{sun2020deepdance, ginosar2019learning} apply adversarial learning in generated dance. \cite{alemi2017groovenet, tang2018dance, yalta2019weakly, zhuang2020towards, kao2020temporally} implement Long short-term memory (LSTM), and \cite{chen2021choreomaster} implement motion graph with learned music/dance embedding through matching approach ~\cite{fan2011example}. More recently, transformer has been applied in dance generation ~\cite{li2020learning, li2021ai, siyao2022bailando, li2022danceformer}. However, the previous methods usually produce temporal freezing frames when generating the long sequences, or are difficult to generalize to in-the-wild music, thereby not satisfiable when directly used for bimodal driven 3D dance generation. In this paper, we design a cross modal transformer which is effective and efficient to integrate text instruction to generate the 3D dance.

\begin{figure*}[t]
	\centering
	\includegraphics[width=\linewidth]{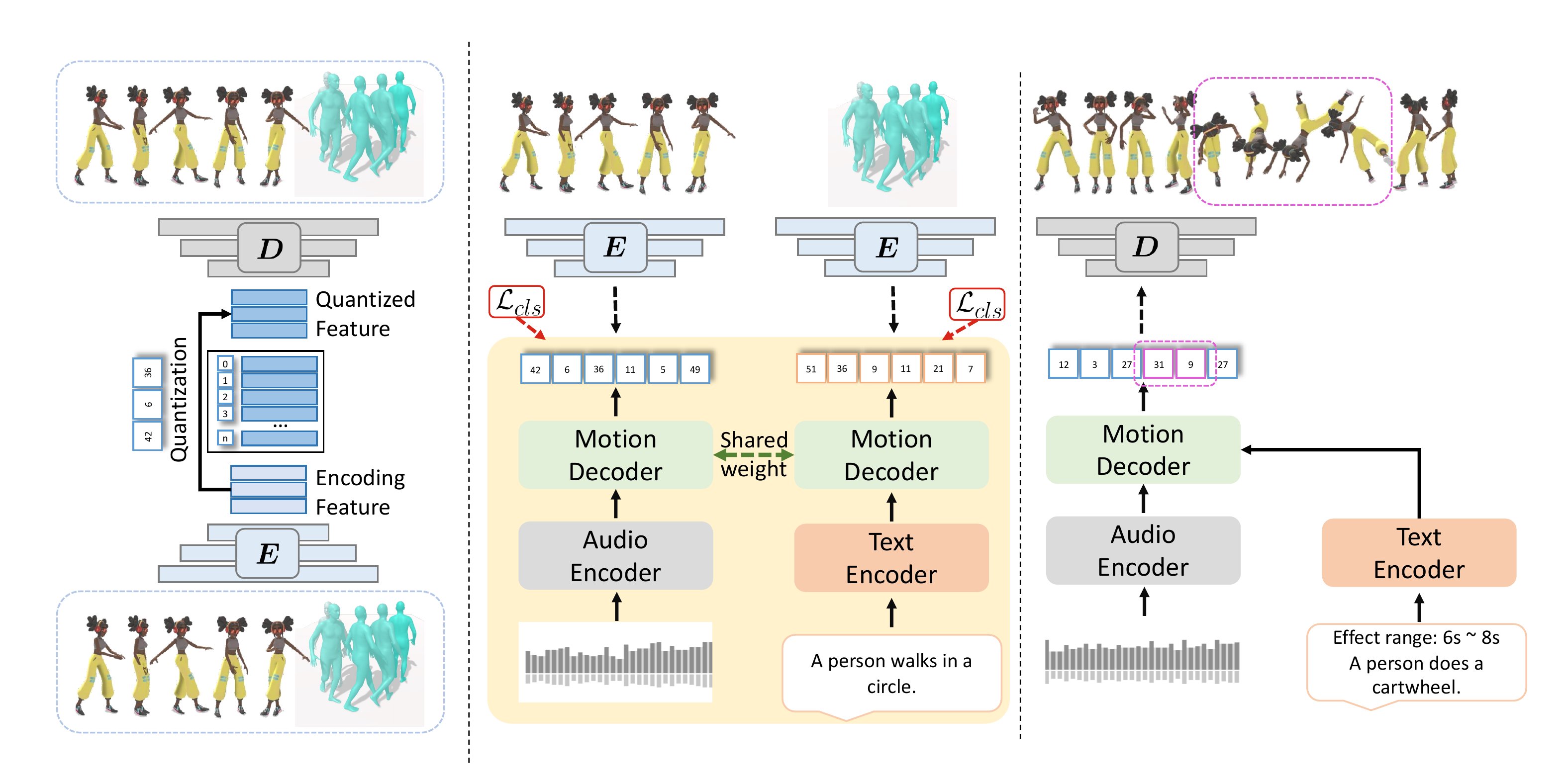}
	\caption{Our proposed pipeline for music-text conditioned 3D dance generation. Three stages from left to right: 3D human motion VQ-VAE, training stage of the cross-modal transformer, and inference stage of our pipeline. In the first stage, a VQ-VAE is trained with both motions from music2dance and text2motion data, which is then used to tokenize all motions. In the second stage, a dual path cross-modal transformer is employed for sequence2sequence translation tasks (\ie, audio to motion tokens, text to motion tokens), with a shared motion decoder. In the third stage, given audio and text inputs, the audio and text encoders first extract the corresponding features, which are then fused (late fusion) in the motion decoder to generate dance condition on both music and text.}
	\label{fig:framework}
\end{figure*}

\subsection{Text to Motion}
In addition to music-driven motion generation, text is also utilized as  instructions to generate motions. Text2motion can be categorized into action label based motion generation and language description based motion generation. The action label based methods ~\cite{guo2020action2motion, petrovich2021action} generate motion conditioned on action labels. Action2Motion ~\cite{guo2020action2motion} implement a GRU-VAE to iteratively generate the next frame based on the action label and previous frames. ACTOR ~\cite{petrovich2021action} implements a transformer-VAE to encode and decode the whole pose sequence in one-shot. Since the action label based methods are restricted in a small set of action labels, language description based methods are proposed for more flexible motion generation. \cite{plappert2018learning, lin2018generating, ahn2018text2action, guo2022generating, guo2022tm2t} formulate the text-to-motion task as a machine translation problem, others ~\cite{yamada2018paired, ahuja2019language2pose, ghosh2021synthesis, petrovich2022temos} learns a joint embedding space of text and motion. Among them, ~\cite{plappert2018learning, lin2018generating, ahn2018text2action, yamada2018paired, ahuja2019language2pose, ghosh2021synthesis} use RNN encoder-decoder  to learn the mapping between text and motion. ~\cite{guo2022tm2t, guo2022generating} use transformer encoder with RNN decoder, and TM2T~\cite{guo2022tm2t}  further introduces motion to text as an inverse alignment for text to motion. TEMOS~\cite{petrovich2022temos} and TEACH~\cite{athanasiou2022teach} use transformer for both encoder and decoder for one-shot generation. Different from these text2motion methods, we focus on how to integrate two modalities (\ie, music and text) together for 3D dance generation. 
\section{Method}\label{sec: method}

\subsection{System Pipeline}
The overall pipeline is shown in Figure \ref{fig:framework}, which consists of three stages. The first stage employs a 3D human motion VQ-VAE to encode both the motions of the music2dance and text2motion datasets to a shared codebook encoded by multiple vectors such that each dance motion can be represented as a discrete motion token. After that, the combinations of music, text, and motion token are fed into a cross-modal transformer for training, which effectively learn how to predict a sequential motion tokens according to the previous those as well as music and text. In the third stage, a random starting motion token is generated and inputted to the cross-modal transformer with the given music and text for the sequential motion token prediction, which will be decoded by the pre-trained 3D human motion VQ-VAE in the first stage to generate 3D dance.

\subsection{3D Human Motion VQ-VAE}
Since there are no paired 3D dance motions conditioned on both music and text, we try to tackle our task with the existing music2dance and text2motion datasets. Instead of directly mixing the motions from both datasets for training, we employ a VQ-VAE to project the motions into a consistent and shared latent space, which is represented with a codebook, as illustrated in Figure \ref{fig:framework}.
To be specific, given a 3D human motion $M \in \mathbb{R}^{T \times d_m}$, where $T$ is the time length and $d_m$ is the dimension of the human motion, an encoder of VQ-VAE that consists of several 1-D convolutional layers projects $M$ to a latent vector $z \in \mathbb{R}^{T' \times d}$, where $T' = \frac{T}{t}$ and $t$ is the time interval for downsampling and $d$ is the dimension of the latent vector. A learnable codebook $e \in \mathbb{R}^{K \times d}$ describes all latent variable features of the whole dataset, where $K$ and $d$ are the length and dimension of the codebook, respectively. A quantized latent vector $z_q \in \mathbb{R}^{T' \times d}$ will record the closest vector from codebook $e$ as follows
\begin{equation}\label{eq: z_q}
	z_{q, i} =  \arg\min_{e_j \in e}{\|z_i - e_j\|} \in \mathbb{R}^{d},
\end{equation}
and the motion token $t_m \in \mathbb{R}^{T' \times 1}$ stores the index of the closest vector
\begin{equation}\label{eq: t_m}
	t_{m, i} =  \arg\min_{j}{\|z_i - e_j\|} \in \mathbb{R}^{1}.
\end{equation}
The quantized $z_q$ will be decoded with stacked convolutional layers to reconstruct the human motion $\hat{M}$, and $t_m$ will be used in the training stage of the cross-modal transformer. 

For the training of 3D human motion VQ-VAE, we follow the strategy in \cite{van2017neural} and the total loss contains a reconstruction loss for human motion regression, a codebook loss for the dictionary learning, and a commitment loss to stabilize the training process: 
\begin{equation}\label{eq: vq_loss}
	\mathcal L_{vq} = \|\hat M  - M \|_1 + \|{\rm sg}[e] - e_q\|^2_2 + \beta \| e - {\rm sg}[{e_{ q}}]\|^2_2,
\end{equation}
where ${\rm sg} [\cdot]$ is `stop gradient' and $\beta$ is the factor term to adjust the weight of the commitment loss. A straight-through estimator is also employed to pass the gradient from the decoder to the encoder in back-propagation. 

As we will show in Sec. \ref{sec: exps}, the motion tokens from both datasets encoded by 3D human motion VQ-VAE fall almost in the shared latent space, which shows the feasibility of using the separate music-conditioned and text-conditioned motion datasets for music-text conditioned 3D dance generation task.

\subsection{Cross-modal Transformer}
The cross-modal transformer contains an audio encoder, a text encoder, and a motion decoder. Since we use two separate datasets for our task, the cross-modal transformer is divided into two branches, where one for music2dance and the other for text2motion. It takes the audio feature, the text feature, and the motion token $t_m$ encoded by 3D human motion VQ-VAE as inputs, and performs the sequence-to-sequence translation task to generate the future motion tokens.

\noindent \textbf{Attention.}
Attention is introduced in Transformer \cite{vaswani2017attention} for natural language processing, and then applied to many domains \cite{Lian_2022_SSF,VanillaNet,ZhangyangGraph1,SGFormerICCV23,fang2023structural,yang2022dery,Xinyin2023structural,MetaFormerBasline,SuchengShunted22}
A $L$-layer transformer typically consists of $L$ transformer blocks, which contains a multi-head self-attention (MSA) and a feed forward network (FFN). Given the input $x \in \mathbb{R}^{N \times c}$, where $N$ is the sequence length and $c$ is the dimension of the input, the transformer block first maps it to keys $K \in \mathbb{R}^{N \times c}$, queries $Q \in \mathbb{R}^{N \times c}$, values $V \in \mathbb{R}^{N \times c}$ with the linear projections, and then a self-attention operation is performed by 
\begin{equation}\label{eq: attention}
	{\rm Attention}(Q, K, V) = {\rm Softmax} (\frac{QK^T}{\sqrt{c}})V.
\end{equation}
The output is followed by a normalization layer and a FFN.

\noindent \textbf{Audio encoder.}
Given a sequence of music, the audio encoder first extracts the raw audio features following Bailando~\cite{siyao2022bailando}, where a public audio processing toolbox, Librosa \cite{jin2017towards}, is employed to obtain the \emph{mel frequency cepstral coefficients (MFCC)}, \emph{MFCC delta}, \emph{constant-Q chromagram}, \emph{tempogram} and \emph{onset strength}. 
After that, an embedded layer followed by several transformer blocks containing self-attention operation \eg, Eq. (\ref{eq: attention}), is used to generate the processed audio features $f_{a} \in \mathbb{R}^{T' \times d}$.

\noindent \textbf{Text encoder.}
Given a text instruction, we first extract the text token with a GloVe \cite{pennington2014glove}, and then an embedded layer followed by several transformer blocks is performed to obtain a processed text feature $f_t \in \mathbb{R}^{n \times d}$, where $n$ is the length of the text feature.

\noindent \textbf{Motion decoder.}
Motion decoder takes the motion tokens encoded by the 3D human motion VQ-VAE as inputs and outputs the future motion tokens, which has the same spirit as the sequence-to-sequence translation task. Since the information after time $t$ is unknown at moment $t$, a masked MSA is first employed to interact with motions at different times. The mask has the shape of a simple upper triangular matrix and is performed in Eq. (\ref{eq: attention}). For the moment $t$, only the motions before moment $t$ are able to perform self-attention operations. Motion decoder also maps the extracted audio feature $f_a$ and text feature $f_t$ via audio encoder and text encoder to $K$ and $Q$, to perform cross-modal attention with motion tokens. To enable the music-text conditional dance generation, (\ie, feature fusion, we apply shared parameters), we use the motion decoder with shared parameters for model efficiency, as shown in Figure \ref{fig:framework}.

\subsection{Details of Music-text Fusion}
\begin{figure}[t]
	\centering
	\includegraphics[width=\linewidth]{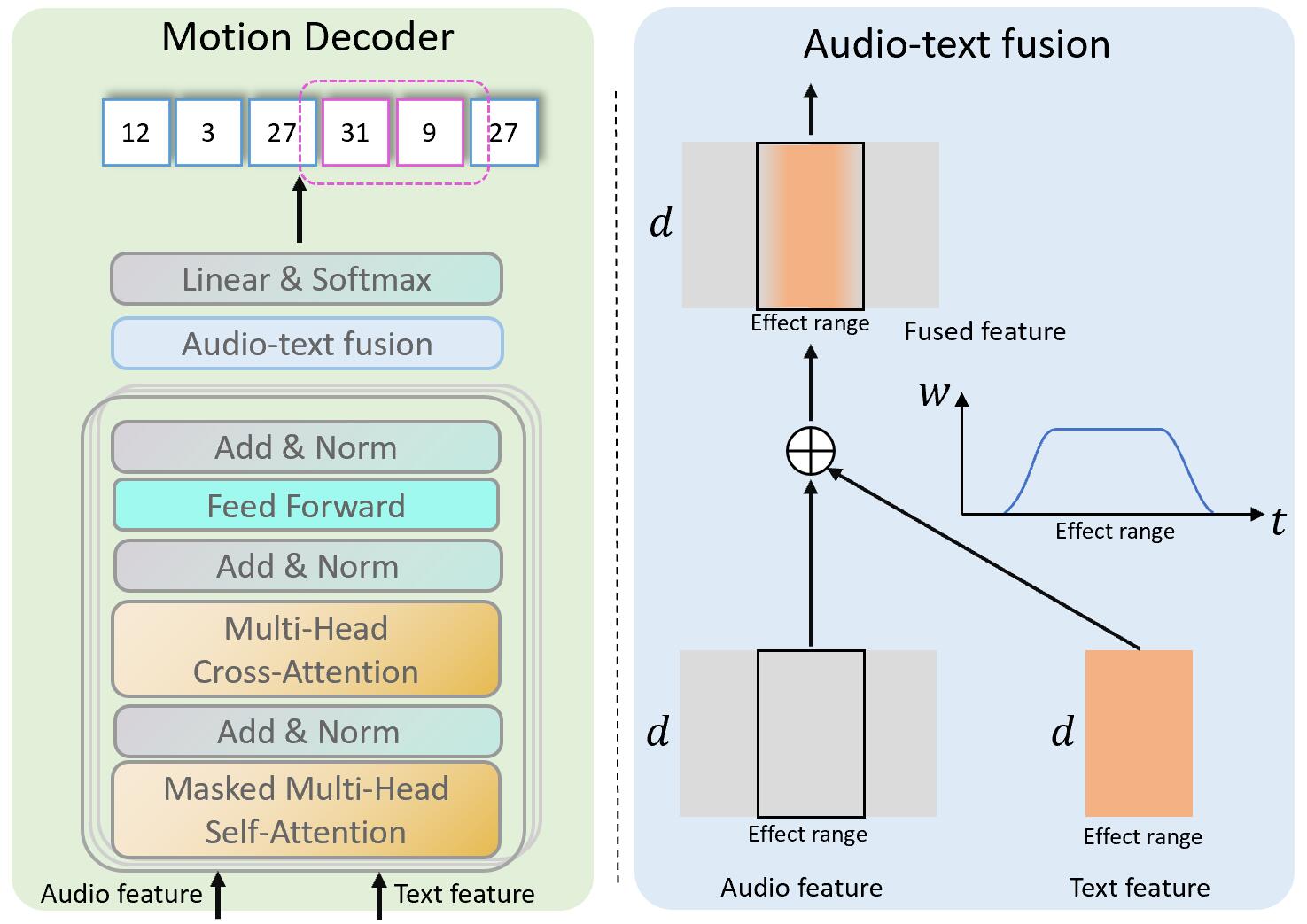}
	\caption{Details of music-text feature fusion: Given the audio and text features from the encoder, the decoder processes them separately in the early layers. Then the audio-text fusion layer is applied by the weighted sum at the effect range with weight cure. Finally, a linear projection layer and a softmax operation are applied to predict the music-text conditioned motion tokens.}
	\label{fig:fusion-detail}
\end{figure}

The detailed architecture is listed in Figure~\ref{fig:fusion-detail}. Given the audio feature and text feature, the motion decoder first processes the past motion tokens with a self-attention layer followed by addition, normalization, a cross-attention with audio and text features separately, addition, normalization, and a feed-forward layer are performed. Such a procedure is repeated $L$ ($L$ = 6) times to build a typical $L$-layer transformer.
Then we adopt a late fusion strategy to perform a weighted sum of features of audio and text at a specific time (\ie, effect range).
The weight curve is shown in Figure~\ref{fig:fusion-detail}, where we slightly increase the weight of the text feature by a half cosine curve until a peak value of 0.8 at the beginning (20\% time of effect range), and decrease it by a half cosine curve at the end (20\% time of effect range). The weight of audio feature $W_{audio}$ is 1 - $W_{text}$ to ensure the feature keeps the same scale.
With the fused feature, a linear projection layer and a softmax operation are applied to predict the music-text conditioned motion tokens, which are then decoded by the decoder of human motion VQ-VAE to obtain the music-text conditioned 3D dance sequences.

\subsection{Training and Inference}
\noindent \textbf{Training.}
To train the 3D human motion VQ-VAE, we crop 64 frames with a sliding window from the original motion sequences as inputs, \ie, $T$ = 64, for both music2dance and text2motion datasets. We use a three-layer encoder and decoder so that the time interval of downsampling $t$ is 8. We randomly sample motions from both datasets and employ the Adam optimizer with a batch size of 128 and a learning rate of $1e^{-4}$ to optimizer our 3D human motion VQ-VAE.

To train the cross-modal transformer, we split two branches into two streams to train the music2dance and text2motion tasks on both datasets. For the music2dance task, we use the same frequency as the motion token to sample the music vector so that the music feature has the same time length as the motion token. We perform the sequence to sequence translation with the motion decoder to obtain the prediction of motion tokens conditioned by the music input. As for the text2motion task, since the text and motion token have different lengths, we use a maximum text length of 84 for text2motion translation with a padding strategy. We employ 6 self-attention layers for the audio encoder, text encoder, and motion decoder with the hidden dimension of 512 and 8 heads. The cross entropy loss is adopted for both music2dance and text2motion tasks
\begin{equation}\label{eq: loss}
	\mathcal L_{cls} = -\frac{1}{m} \sum_{i=0}^{m-1} \sum_{j = 0}^{C-1} y_{ij} \log \hat{y_{ij}},
\end{equation}
where $\hat{y}$ is the prediction of motion token and $y$ is the ground-truth. $m$ is the length of motion tokens and $C$ is the number of classes of motion tokens, \ie, the length of codebook of 3D human motion VQ-VAE. Both tasks are simultaneously optimized with a batch size of 64 and a learning rate of $1e^{-4}$.

\noindent \textbf{Inference.}
Our aim is to generate 3D dance with music-text integration. Therefore, at the inference stage, we first fuse the music feature and text feature extracted by the audio encoder and text encoder with a weighted sum. Specifically, given an audio feature with a time length of $T'$ and a text feature with the length of $n$, we first feed them into the motion decoder for future motion token prediction. Then we adopt the late fusion strategy to have a weighted sum of the generated motion tokens from both audio and text features at the duration of the integrated text, followed by a linear projection layer and a softmax operation, to obtain the combined motion tokens. 
Finally, the combined motion tokens are decoded into a 3D dance sequence after going through the decoder of our 3D human motion VQ-VAE model.

\section{Experiments}\label{sec: exps}
\subsection{Experimental Settings}
\noindent \textbf{Datasets.} 
Since there are no paired 3D dance motions conditioned on both music and text for our task, we combine the existing music2dance and text2motion datasets. 
For the music2dance dataset, we employ the AIST++ dataset~\cite{li2021ai}. 
For the text2motion dataset, we employ the HumanML3D~\cite{guo2022generating} dataset. 
Additionally, we also collect a new in-the-wild music dataset to evaluate the generalization ability of our method, which contains 82 wild music clips from Youtube. 
The total of 53 minutes duration (8x larger size than the AIST++ test set) and various styles and content of music, which faithfully lies out of the distribution of AIST++ (explained in supplementary material). The evaluation on our dataset better reflects the generalization ability of the models in real-world scenarios.

\noindent \textbf{Baselines.} Since we are the first to propose this bimodality driven 3D dance generation task, there are no existing methods for comparisons. We implement a {traditional motion editing algorithm}, slerp~\cite{shoemake1985animating}, which aligns the last generated frame of the dance with the first frame of the motion by text with a transition window of 10 frames, and applies spherical interpolation of quaternion to fill in the transition in between. 
To validate the effectiveness of our method on music2dance task, we compare our method to the previous music2dance works~\cite{li2020learning, zhuang2022music2dance, huang2020dance, li2021ai, siyao2022bailando}. 
To validate the effectiveness of our method on text2motion task, we compare our method to the previous text2motion works~\cite{lin2018generating, ahuja2019language2pose, bhattacharya2021text2gestures, ghosh2021synthesis, tulyakov2018mocogan, lee2019dancing, guo2022tm2t}.

\begin{table*}[t]
\centering
\small
\setlength{\tabcolsep}{5pt}
{
\begin{tabular}{l l c c c c c c c c}
\toprule
    \multirow{2}{*}{Data} & \multirow{2}{*}{Method} &  \multicolumn{2}{c}{ Motion Quality} & \multicolumn{2}{c}{ Motion Diversity } & \multicolumn{2}{c}{Freezing} & \multirow{2}{*}{Beat Align Score $\uparrow$} & \multicolumn{1}{c}{ User Study}      \\
    \cmidrule(r){3-4} \cmidrule(r){5-6} \cmidrule(r){7-8}    \cmidrule(r){10-10}   
    ~ & ~ & FID$_k\downarrow$ & FID$_g^{\dag}\downarrow$ &  Div$_k\uparrow$ &  Div$_g^{\dag}\uparrow$ &  PFF$\downarrow$ &  AUC$_f\downarrow$ &  ~  & Our Method Wins \\
    \midrule
    \multirow{8}{*}{\makecell[c]{AIST\\++}} & Ground-truth & 17.10 & 10.60 & 8.19 & 7.45 & 0.00 & 0.00 & 0.2374 &  41.9\%   \\
    \cmidrule(r){2-10}
    ~ & Li \etal~\cite{li2020learning} & 86.43 & 43.46 & 6.85$^*$ & 3.32 & \textbf{0.00} & \textbf{0.00}  & 0.1607 & 98.3\%  \\
    ~ & DanceNet~\cite{zhuang2022music2dance} & 69.18 & 25.49  & 2.86 &  2.85 & \textbf{0.00} & \underline{0.98}  & 0.1430 & 87.2\%  \\
    ~ & DanceRevolution~\cite{huang2020dance} & 73.42 & 25.92 & 3.52 & 4.87 & \underline{11.01} & 12.22  & 0.1950 & 70.5\%  \\
    ~ & FACT~\cite{li2021ai} & {35.35} & {22.11} & {5.94} & {6.18} & 25.29 & 21.59  & \underline{0.2209} & 83.3\%  \\
    ~ & Bailando~\cite{siyao2022bailando}  &  {28.16} &  \underline{9.62} &  \underline{7.83} &  \underline{6.34} & 14.91 & 13.25  & \textbf{0.2332} & 65.0\%  \\
    \cmidrule(r){2-10}
    ~ & ours (only dance data) & \underline{23.94} &  \textbf{9.53} & 7.69 & 4.53 & \textbf{0.00} & \textbf{0.00}  & 0.2127 & - \\
    ~ & ours & \textbf{19.01} &  {20.09} & \textbf{9.45} & \textbf{6.36} & \textbf{0.00} & \textbf{0.00}  & 0.2049 & --  \\ 
    
    \midrule
    \multirow{4}{*}{\makecell[c]{Wild\\Audio}} & FACT~\cite{li2021ai} & 70.36 & {20.23} & {7.33} & \textbf{6.34} & 32.64 & 27.21 & \underline{0.2211} & 89.6\%  \\
    ~ & Bailando~\cite{siyao2022bailando}  &  {50.56} &  22.55 &  3.80 & \underline{6.04} & {17.55} & {14.14}  & 0.2166 & 70.8\%  \\
    \cmidrule(r){2-10}
    ~ & ours (only dance data) & \underline{43.85} &  \textbf{13.08} & \textbf{8.52} & 4.76 & \underline{0.78} & \underline{0.76}  & 0.1998 & --  \\ 
    ~ & ours & \textbf{27.65} &  \underline{20.34} & \underline{7.88} & 5.27 & \textbf{0.12} & \textbf{0.11}  & \textbf{0.2290} & --   \\
    \bottomrule
\end{tabular}
}
\caption{Music Conditioned Dance Generation: quantitative results on AIST++ and Wild Audio test set. The best and runner-up values are bold and underlined, respectively.
Among compared methods, ``Li \etal'', DanceNet and FACT are multiplexing the same results of AIST++ benchmark~\cite{li2020learning},  while DanceRevolution~\cite{huang2020dance} is followed from Bailando~\cite{siyao2022bailando}.
$\dag$ FID$_k$ and DIV$_k$ are fetched from \cite{li2021ai} while FID$_g$ and DIV$_g$ are  fetched from \cite{siyao2022bailando}.
}
\label{table:music2dance}
\end{table*}

\noindent \textbf{Evaluation metrics.} 
We adopt the same evaluation settings as suggested by FACT \cite{li2021ai} and Bailando \cite{siyao2022bailando} to evaluate the dance generation quality,
including Fréchet Inception Distances (FID) \cite{heusel2017gans}, Diversity, and Beat Align Score. 
For the text2motion quality, we adopt the same evaluation settings as suggested by TM2T~\cite{guo2022tm2t}, including R-precision, Multimodal-Dist, FID, Diversity, and MultiModality.

In addition, we also propose two new evaluation metrics, Motion Prediction Distance (MPD) and Freezing Score. The former reflects the coherence of frames at the time of integrating text, and the latter reacts to the percentage of
temporal freezing frames in the generated dance.
For the Freezing Score, we introduce the Percentage of Freezing Frame (PFF) and AUC$_f$, where the PFF is defined by measuring the percentage of frozen frames with two criteria: 1) the maximum joint velocity blew a threshold (0.015m/s). 2) its duration exceeds a certain period (3s). The AUC$_f$ is defined by area under curve of PFF in the threshold range from 0 to 0.03. As for MPD, it is defined as 
\begin{equation}
    {\rm MPD} = \min_{i}\|f_i(M_{t_0 \rightarrow t_1}) - M_{t_1 \rightarrow t_2} \|_2,
\end{equation}
where $M$ is the dance motions, $f$ is a motion prediction model, $f_i(M_{t_0 \rightarrow t_1})$ is the i-th predicted possible future motion, and $t_0$, $t_1$, and $t_2$ are the timestamps. It means that the model predicts various potential possible motion from $t_1$ to $t_2$ with the motion from $t_0$ to $t_1$. If this distance is small (\ie, the generated dance $M_{t_1 \rightarrow t_2}$ lays in the possible future), then the generated dance motion $M_{t_1 \rightarrow t_2}$ is more coherent at the time of integrating text. We adopt the motion prediction model DLow~\cite{yuan2020dlow} in this evaluation metric as it is designed for diverse potential furture motion prediction.

\noindent \textbf{Perceptual Evaluation.} Besides the above metric measuring, we also conduct extensive user studies on Amazon Mechanical Turk (AMT) to perceptually evaluate the visual effects of our generated 3D dance results. Particularly, given each pair of dance movements sampled from our method and others with the same music clip, we request 3 distinct users on AMT to present their preference regarding the music-dance alignment, motion realism, and mobility. We further set the bar of involved users that only the ones with Master recognition who also have finished more than 1,000 tasks with over 98\% approve rate are considered. Overall, there are 55 users employed in our user studies that come from various regions, ages, races and gender. The results of the user study are more representative to show the effect of the generated dance in practice. 

\begin{table}
\centering
\small
\setlength{\tabcolsep}{4.5pt}
{
\begin{tabular}{l l c c c c c c c}
\toprule
    \multirow{2}{*}{Data} & \multirow{2}{*}{Method} & \multicolumn{6}{c}{ Motion Prediction Distance} & \multicolumn{1}{c}{ User Study}        \\
    \cmidrule(r){3-8} \cmidrule(r){9-9} 
    ~ & ~ & ft=10 & ft=20 & ft=30 & Our Method Wins \\
    \midrule
    \multirow{5}{*}{\makecell[c]{AIST\\++}} & GT & 0.048 & 0.072 & 0.088 & 70.0\%  \\
    \cmidrule(r){2-9}
    ~ & a2d & \underline{0.052} & \underline{0.084} & \underline{0.108} & 66.6\%  \\
    ~ & slerp~\cite{shoemake1985animating} & 0.088 & 0.122 & 0.135 & 73.3\%  \\
    \cmidrule(r){2-9}
    ~ & ours & \textbf{0.049} & \textbf{0.080} & \textbf{0.102} & -  \\
    \midrule
    \multirow{3}{*}{\makecell[c]{Wild\\Audio}} & a2d & \underline{0.052} & \underline{0.088} & \underline{0.113} & 60.0\%  \\
    ~ & slerp~\cite{shoemake1985animating} & 0.104 & 0.148 & 0.171 &  70.0\% \\
    \cmidrule(r){2-9}
    ~ & ours & \textbf{0.048} & \textbf{0.079} & \textbf{0.107} & -  \\
    \bottomrule
\end{tabular}
}
\caption{Music-text conditioned Dance Generation: quantitative results on AIST++ and wild audio test set. The best and runner-up values are bold and underlined, respectively.
}
\label{table:tm2d}
\end{table}

\subsection{Evaluation on Music-text Conditioned Dance Generation}
We validate the results of music-text conditioned dance generation on the test set of AIST++ and our dataset, as illustrated in Table \ref{table:tm2d}. Here we mainly evaluate the coherence of generated dance with MPD metric at the time where music meets text, \ie, the time point where text starts to take effect. We measure the coherence of pure music2dance generation as a baseline named a2d to reflect the influence of with/without text instruction. We use the past 25 motion frames to predict the future 30 frames, and calculate the MPD from future frame (ft) = 10 to ft = 30, respectively. 
Our method consistently outperforms slerp \cite{shoemake1985animating} and a2d baseline in both datsets, and gains a similar result compared to the ground-truth shown in Table \ref{table:tm2d}, which verifies the naturalness of our generated motion.
More importantly, user study experiments show that our method generates more realistic 3D dance compared to dance generation only conditioned music, and 70.0\% Win rate even compared to the ground-truth shown in Table \ref{table:tm2d}. The results of our user study indicate that the music-text conditioned dance generation received high ratings from participants, highlighting the importance of considering audience preferences in evaluating dance quality. 

\subsection{Evaluation on Music Conditioned Dance Generation}
To validate the effectiveness of our architecture, we quantitatively compare our method with state-of-the-art those for music conditioned 3D dance generation. The results on the test set of AIST++ are shown in Table \ref{table:music2dance}. We can find that our method outperforms the previous ones in terms of motion diversity (Div$_k$ and Div$_g$), while the performance of the motion quality (FID$_g$) and beat align score is inferior under the condition that we do not use seed motion from ground-truth compared with FACT \cite{li2021ai} and Bailando~\cite{siyao2022bailando}. 

The existing evaluation metrics are not sufficient to reflect the quality of generated dance in practice, which motivates us to propose a PFF and AUC$_f$ to evaluate the percentage of freezing frames. It is worth noting that our architecture outperforms FACT \cite{li2021ai} and Bailando~\cite{siyao2022bailando} in terms of PFF and AUC$_f$, which shows that our method rarely generates frozen frames. Meanwhile, Li \etal~\cite{li2020learning} and DanceNet~\cite{zhuang2022music2dance} also gain nearly zero freezing in PFF and AUC$_f$. The reason is that the generated dances of Li \etal~\cite{li2020learning} are highly jittery making its velocity variation extremely high, which is also reported in \cite{li2021ai, siyao2022bailando}, and leads to the non-freezing issue. And for DanceNet~\cite{zhuang2022music2dance}, it generates dance in a repeat motion pattern, nearly zero freezing but with low diversity. 
Furthermore, the results of user study show that the generated 3D dance is more visually realistic compared to other methods. Even in comparison to the ground truth, 41.9\% of our generated dance is voted as the better in average. 
We also report the result trained with dance-only data, which shows comparable performance.

In addition to the evaluation on the test set of AIST++, we also show the experimental results on our in-the-wild dataset in Table \ref{table:music2dance}. Our method outperforms FACT \cite{li2021ai} and Bailando \cite{siyao2022bailando} in almost all metrics except for FID$_g$ Div$_g$. We can observe that both FACT \cite{li2021ai} and Bailando \cite{siyao2022bailando} shows a large performance drop in terms of FID$_k$, while ours maintain a small change. This is because FACT \cite{li2021ai} and Bailando \cite{siyao2022bailando} requires seed motion, however, there is no ground-truth for in-the-wild scenario. With random sampled seed motion or token, their methods are not adapted well for the in-the-wild scenarios. One can also notice that there exists freezing in our method on the in-the-wild dataset but the percentage of frozen frames is zero on the test set of AIST++, which results from the different distributions of AIST++ and our dataset. It also shows that our dataset is more challenging. 
We also report the result trained with dance-only data, which shows the advantage of our mix-training strategy in the wild scenario.

\subsection{Evaluation on Text Conditioned Motion Generation.}
\label{sec:exp-t2m}
We evaluate our text2motion approach in two different scenarios: inference with text only (t2m), and inference with text and music feature fusion. Table~\ref{table:text2motion} shows the results. In the text-only setting, our approach achieves comparable performance with TM2T baseline(T)~\cite{guo2022tm2t}, which demonstrates that the mixed data/tasks training does not affect the quality of text2motion generation. We then apply our late fusion method by randomly sampling a music clip (from the whole AIST++ test set) and a text (from the whole t2m test set), and evaluate the generated dance clips following the same protocol as TM2T~\cite{guo2022tm2t}. As Table~\ref{table:text2motion} indicates, the text-dance consistency remains acceptable with a late fusion ratio of 0.8. For more details on the relation between the late fusion rate and text2motion result, please refer to the supplementary material.

\begin{table}[t]
\renewcommand{\et}[2]{#1}
\centering
\setlength{\tabcolsep}{0.5pt}
\scriptsize
{
\begin{tabular}{l c c c c c c c}
\toprule
    Methods  & R Precision$\uparrow$ &  FID$\downarrow$ &  MM Dist$\downarrow$ &  Diversity$\rightarrow$ &  MModality$\uparrow$  \\
    \midrule
    {Real motions} & {0.511} & {0.002} & {2.974} & {9.503} & -  \\
    \midrule
    Seq2Seq~\cite{lin2018generating} & \et{0.180}{.002} & \et{11.75}{.035} & \et{5.529}{.007} & \et{6.223}{.061}  & -  \\

    Language2Pose~\cite{ahuja2019language2pose} & \et{0.246}{.002} & \et{11.02}{.046} & \et{5.296}{.008} & \et{7.676}{.058} & -  \\
    
    Text2Gesture~\cite{bhattacharya2021text2gestures} & \et{0.165}{.001} & \et{5.012}{.030} & \et{6.030}{.008} & \et{6.409}{.071} & -  \\
    
    Hier~\cite{ghosh2021synthesis} & \et{0.301}{.002} & \et{6.532}{.024} & \et{5.012}{.018} & \et{8.332}{.042} & -  \\

    MoCoGAN~\cite{tulyakov2018mocogan} & \et{0.037}{.000} & \et{94.41}{.021} & \et{9.643}{.006} & \et{0.462}{.008} & \et{0.019}{.000}  \\

    Dance2Music~\cite{lee2019dancing} & \et{0.033}{.000}& \et{66.98}{.016} & \et{8.116}{.006} & \et{0.725}{.011} & \et{0.043}{.001}  \\
    
    TM2T baseline(T)~\cite{guo2022tm2t} & \et{\underline{0.351}}{.003} & \et{1.669}{.025} & \et{\underline{4.046}}{.018} & \et{\underline{9.632}}{.072} & \et{\underline{4.352}}{.149}  \\
    
    TM2T~\cite{guo2022tm2t} & \et{\textbf{0.424}}{.003} & \et{1.501}{.017} & \et{\textbf{3.467}}{.011} & \et{8.589}{.076} & \et{2.424}{.093}  \\

    \midrule
    TM2D (t2m) & \et{0.319}{.003} & \et{\textbf{1.021}}{.022} & \et{{4.098}}{.014} & \et{\textbf{9.513}}{.101} & \et{4.139}{.116}  \\
    TM2D (LFR 0.8) & \et{0.300}{.0027} & \et{\underline{1.105}}{0.0569} & \et{{4.307}}{.0170} & \et{8.887}{.0760} & \et{\textbf{4.443}}{.1295}  \\
    \bottomrule
\end{tabular}
}
\caption{The evaluation of text2motion on the HumanML3D dataset, where the results are averaged from 20 runs. 
}
\label{table:text2motion}
\end{table}

\subsection{Qualitative Results}
\begin{figure}[t]
	\centering
	\includegraphics[width=\linewidth]{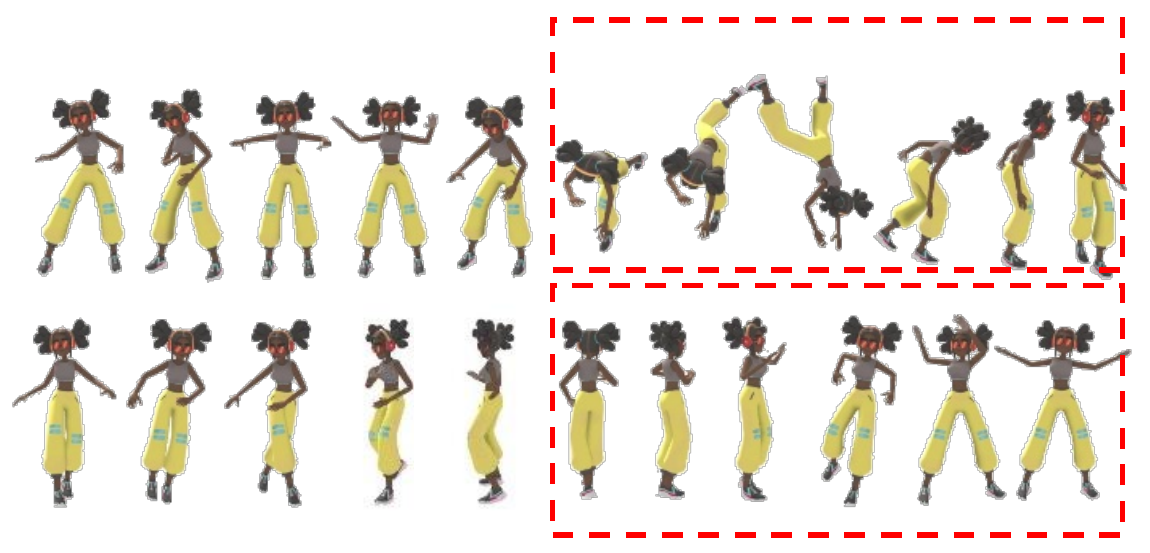}
    \caption{\textbf{Same music, different actions.} These two generated dance share the same music and effect range (from 6s to 8s), with different text instructions: ``A person does a cartwheel'' (top), ``A person is spinning with arms spread out'' (bottom).}
	\label{fig:samemusicdiffaction}
\end{figure}

\begin{figure}[t]
	\centering
	\includegraphics[width=\linewidth]{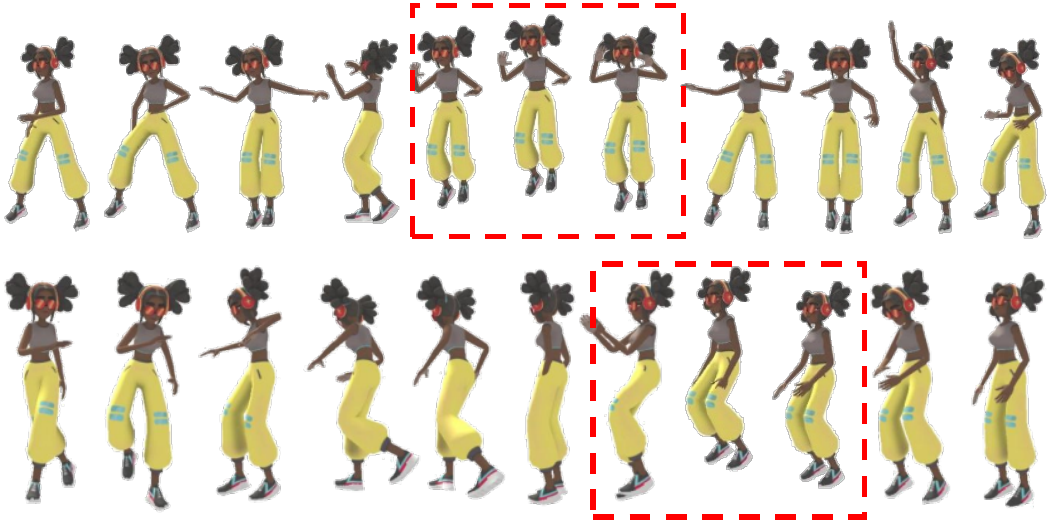}
    \caption{\textbf{Same action, different start time points.} These two generated dance share the same music and text instruction (``A person jumps up and down'') but different effect ranges: from 6 to 8s (top), from 7 to 9s (bottom).}
	\label{fig:sameactdiffstart}
\end{figure}

\begin{figure}[t]
	\centering
	\includegraphics[width=\linewidth]{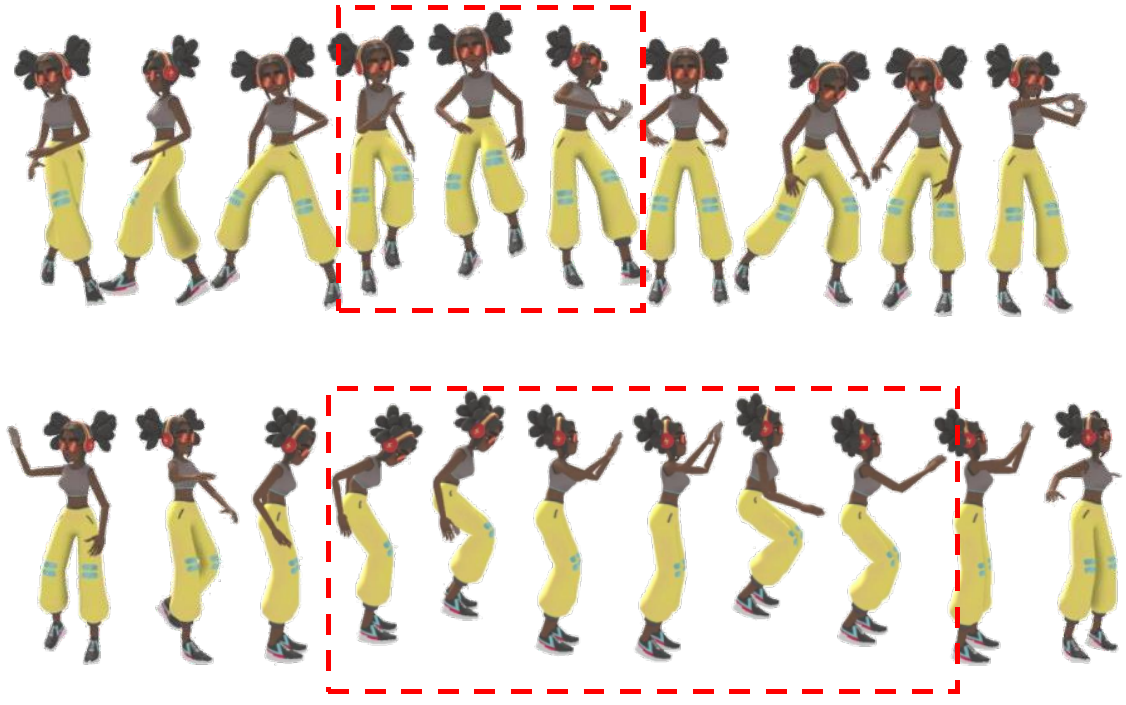}
  \caption{\textbf{Same action, different durations.} These two generated dance share the same music and text instruction (``A person is keeping jumping'') and action start time point (\ie, 6s) but different effect duration: 2s (top), 3s (bottom, \ie, jumped twice).}
	\label{fig:sameactdiffduration}
\end{figure}

We also show the qualitative results of our method for music-text conditioned 3D dance generation in Figure \ref{fig:samemusicdiffaction}, Figure \ref{fig:sameactdiffstart} and Figure \ref{fig:sameactdiffduration}. Specifically, Figure \ref{fig:samemusicdiffaction} shows the generated 3D dance with the same music sequence but with different text instructions. Figure \ref{fig:sameactdiffstart} shows the generated 3D dance with the same text instruction but with a different start time. Figure \ref{fig:sameactdiffduration} shows the generated 3D dance with the same text instruction but with different durations. As one can see, our approach maintains plausible visual results according to the text instructions for all three cases, which confirms that our approach is more flexible.  

\subsection{The Impact of Mixed Data}
\begin{figure}[h]
    \centering
    \subfloat[Before VQ-VAE]{
    \begin{minipage}{0.5\linewidth}
        \centering
        \includegraphics[width=\linewidth]{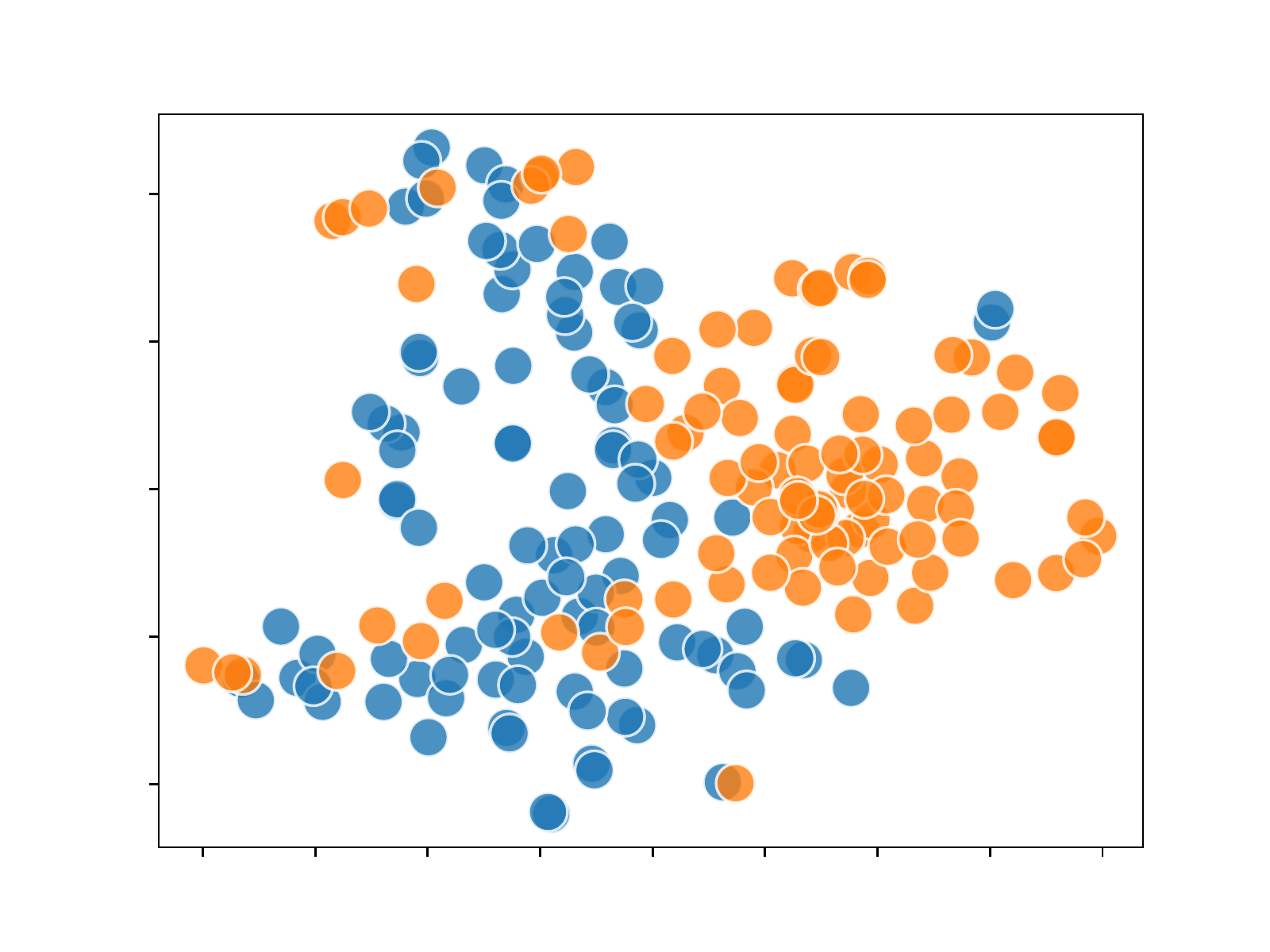}
    \end{minipage}}
    \subfloat[After VQ-VAE]{
    \begin{minipage}{0.5\linewidth}
        \centering
        \includegraphics[width=\linewidth]{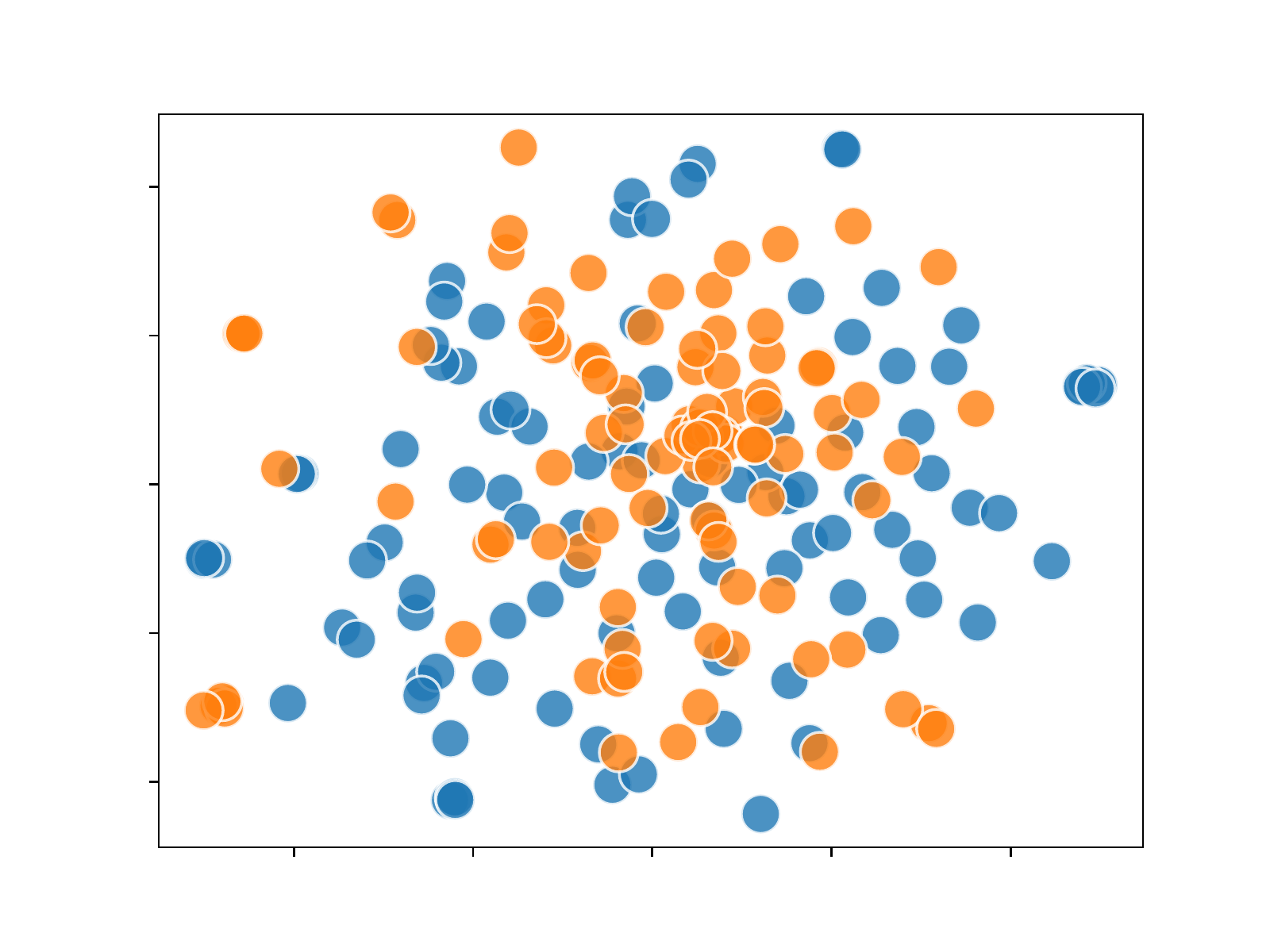}
    \end{minipage}}
    \caption{The t-SNE visualizations of motions before and after VQ-VAE (orange: motions in AIST++, blue: motions in HumanML3D). 
    }
    \label{fig:tsne-motion}
\end{figure}

We employ two datasets (AIST++ and HumanML3D) to train our 3D human motion VQ-VAE so that the motion tokens drop in the same latent space. To empirically show this point, we conduct experiments to count the number of shared motion tokens. For a trained 3D human motion VQ-VAE with AIST++ and HumanML3D, there are 855 vectors and 912 vectors from codebook used to construct the motion token in AIST++ and HumanML3D. The total number of vectors contained in the codebook is 1024. among them, 846 vectors (98.9\% in AIST++ and 92.8\% in HumanML3D) are shared to generate the motion tokens, which shows the feasibility of using the separate music-conditioned and text-conditioned motion datasets for music-text conditioned 3D dance generation task. 
We further verify the latent space of the motions by VQ-VAE, and we perform t-SNE visualizations of the raw motion distribution (before VQ-VAE) and feature distribution encoded by VQ-VAE (after VQ-VAE) from a2d motion and t2m motion in Figure~\ref{fig:tsne-motion}. Two motion datasets are mixed successfully, which provides the potential integration of bi-modality dance generation. Refer to the supplementary materials for more details.

\subsection{Model Efficiency}
We also show the efficiency of our architecture in Table \ref{table: model efficiency}. In the inference stage, the complete model parameters and the inference time for 1s music are compared. Our architecture requires only about half model parameters and inference time compared to Bailando \cite{siyao2022bailando}, which attributes that we do not need to encode separate features for the upper and lower half body. Compared to FACT \cite{li2021ai}, our architectures significantly reduce the inference time. The main reason is that FACT employs two transformers for the music, dance motion, and a cross-modal transformer, but we formulate it as the standard sequence-to-sequence translation task with one transformer only, which establishes the advantage of our architecture. 

\begin{table}
\centering
{
\begin{tabular}{l c c }
\toprule
    Method & Parameters & Inference time  \\
    \midrule
    FACT \cite{li2021ai} & 120M  & 8.300s/(1s music) \\
    Bailando \cite{siyao2022bailando} & 152M & 0.236s/(1s music)   \\
    \midrule
    ours & 72M & 0.143s/(1s music)  \\
    \bottomrule
\end{tabular}
}
\caption{Comparisons of parameters and inference time.}
\label{table: model efficiency}
\end{table}
\section{Conclusion}\label{sec: conclusion}
In this paper, a novel task that simultaneously integrates both music and text instruction for 3D dance generation is proposed. Due to the lack of the paired 3D dance motions conditioned on both music and text, we resort to two existing datasets, \ie, music2dance and text2motion, to perform this task and employ a 3D human motion VQ-VAE to project the motions of the two datasets into a shared latent space so that the two datasets with different distributions can be effectively mixed for training. Moreover, we also propose a cross-modal transformer architecture to generate 3D dance without degrading the performance of music-conditioned dance generation. Two new evaluation metrics, MPD and FS, are proposed to reflect the quality of generated motion.
Extensive experiments show that our method can generate dance motion that matches both music and text in a realistic and coherent way while maintaining comparable performance on two single modalities.

\noindent \textbf{Acknowledgement.} 
This project is supported by 
the Ministry of Education
Singapore under its Academic Research Fund Tier 2 (Award Number: MOE-T2EP20122-0006),
and 
Singapore National Research Foundation (``CogniVision – Energy-autonomous always-on cognitive and attentive cameras for distributed real-time vision with milliwatt power consumption'' grant NRF-CRP20-2017-0003) – www.green-ic.org/CogniVision.

{\small
\bibliographystyle{ieee_fullname}
\bibliography{egbib}
}

\clearpage
\appendix
\noindent\textbf{\large APPENDIX: \\}
\begin{abstract}
This supplementary material provides more details on the following aspects of our study: i) The dataset we used; ii) The evaluation metrics we employed; iii) The impact of mixed data for shared motion token space; iv) The collected data distribution; v) The effect of music-text fusion weight; vi) The reason why our dance has less freeze issue; vii) More visualizations of our results.
\end{abstract}

\section{Detail of Dataset}
For the music2dance dataset, we employ the AIST++ dataset~\cite{li2021ai}, which contains 30 subjects and 10 dance genres. There are 992 pieces of 3D human pose sequence, of which 952 are used for training and the rest are used for evaluation. 

For the text2motion dataset, we employ the HumanML3D~\cite{guo2022generating} dataset, which is a large-scale 3D human motion dataset that covers a broad range of human actions such as locomotion, sports, and dancing. It consists of 14,616 motions and 44,970 text descriptions. Each motion clip comes with at least 3 descriptions. For the joint training of both datasets, we sample the motions with 60 frames per second (FPS) to keep the time consistency with the AIST++ dataset, resulting in duration ranges from 2 to 10 seconds.

To evaluate the generalization ability of our method, we also collected a new dataset of music clips from YouTube that are not included in AIST++. This dataset consists of 82 clips with a total duration of 53 minutes, which is eight times larger than the AIST++ test set. The clips cover various styles and content of music, which are out of the distribution of AIST++.
In detail, our data are popular music collected from YouTube, which covers a variety of styles such as Glitch hop, Electro house, rock, future bass, indie pop, and R\&B. By contrast, AIST++ uses pure dance music from Old School (Break, Pop, Lock, and Waack) and New School (Middle Hip-hop, LA-style Hip-hop, House, Krump, Street Jazz, and Ballet Jazz) genres.
Additionally, we selected in-the-wild music based on popularity, such as "Faded," "Beat It," "Coincidence," "Baby," "Poker Face," "Despacito," "Panama," "Love Story," and others, with millions of plays.
Furthermore, we provide a t-SNE feature distribution diagram (Figure~\ref{fig:tsne-audio}) to demonstrate the diversity and distinctiveness of our dataset compared to AIST++.

\section{Evaluation Metrics}
We follow FACT \cite{li2021ai} and Bailando \cite{siyao2022bailando} to quantitatively measure the quality of generated dances, the diversity of motions and the beat alignment of the music and the generated motions. 
In concrete, for the dance quality, 
we calculate the Fréchet Inception Distances (FID) \cite{heusel2017gans} between the generated 3D dance and all motions of the AIST++ dataset on kinetic features \cite{onuma2008fmdistance} (denoted as `$k$') and geometric features \cite{muller2005efficient} (denoted as `$g$') extracted by \cite{gopinath2020fairmotion} to measure the quality of generated dances. We also follow \cite{li2021ai} to calculate the average feature distance of generated motion to measure the diversity of motions. The average distance between the music beat and its closest dance beat is defined as the Beat Align Score as follows
\begin{equation}
    \frac{1}{|B^m|}\sum_{b^m \in B^m}\exp{\left\{-\frac{\min_{b^d \in B^d} {\|b^d - b^m\|^2}}{2\sigma^2}\right\}},
\end{equation}
where $B^d$ and $B^m$ are the dance beats and music beats, respectively. $\sigma$ is a normalized parameter that we set to be $\sigma = 3$ in our experiments.

For the text2motion quality,
we follow the same setting suggested by TM2T~\cite{guo2022tm2t}:
R-precision and Multimodal-Dist quantify the relevancy between the generated motions and the input prompts; 
FID computes the distance between the generated and ground truth distributions (in latent space); 
Diversity evaluates the variation of the generated motions; 
and MultiModality estimates the variance for a single prompt

We also introduce two new evaluation metrics: Percentage of Freezing Frame (PFF) and Motion Prediction Distance (MPD). PFF measures the degree of freezing in the generated dance, while MPD assesses the coherence of frames when text is integrated.

\begin{figure}[t]
    \centering
    \subfloat[Difference in FID$_k$.]{
    \begin{minipage}{0.5\linewidth}
        \centering
        \includegraphics[width=\linewidth]{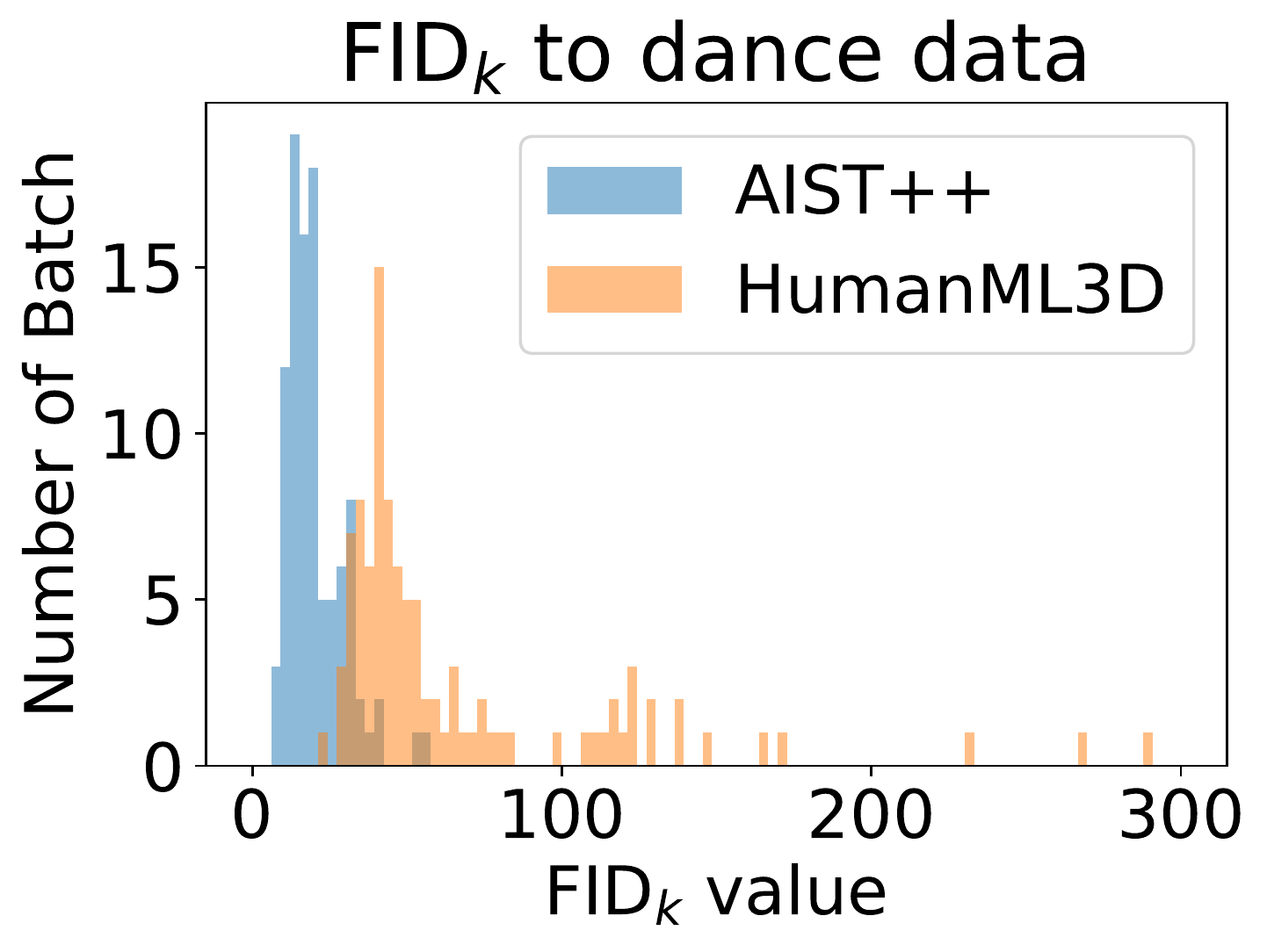}
    \end{minipage}}
    \subfloat[Difference in FID$_g$.]{
    \begin{minipage}{0.5\linewidth}
        \centering
        \includegraphics[width=\linewidth]{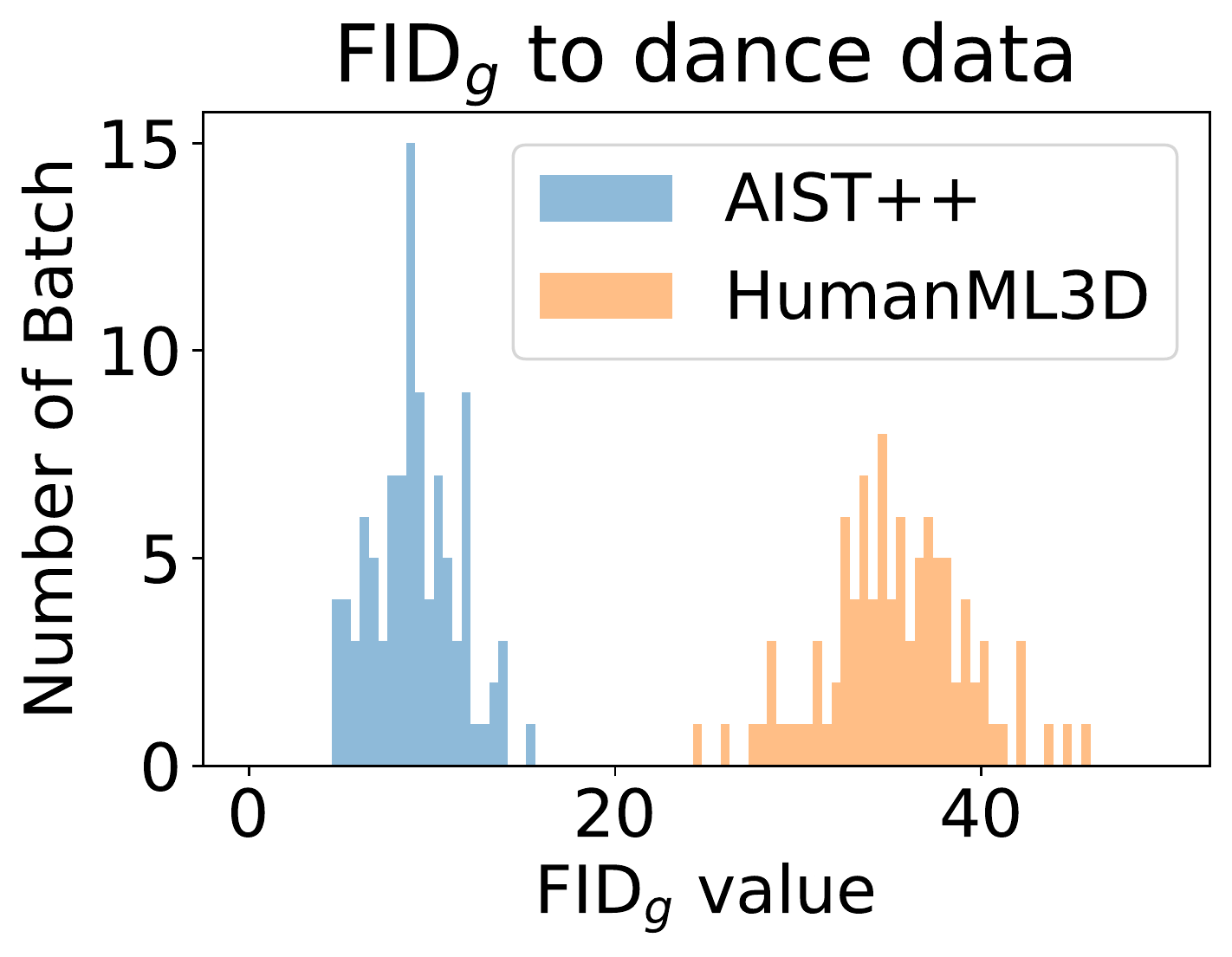}
    \end{minipage}}
    \caption{FID$_k$ and FID$_g$ with difference batches in Experiment A.}
    \label{fig:fid-diff}
\end{figure}

\begin{figure*}[t]
    \centering
    \subfloat[token-of-AIST++]{
    \begin{minipage}{0.24\linewidth}
        \centering
        \includegraphics[width=\linewidth]{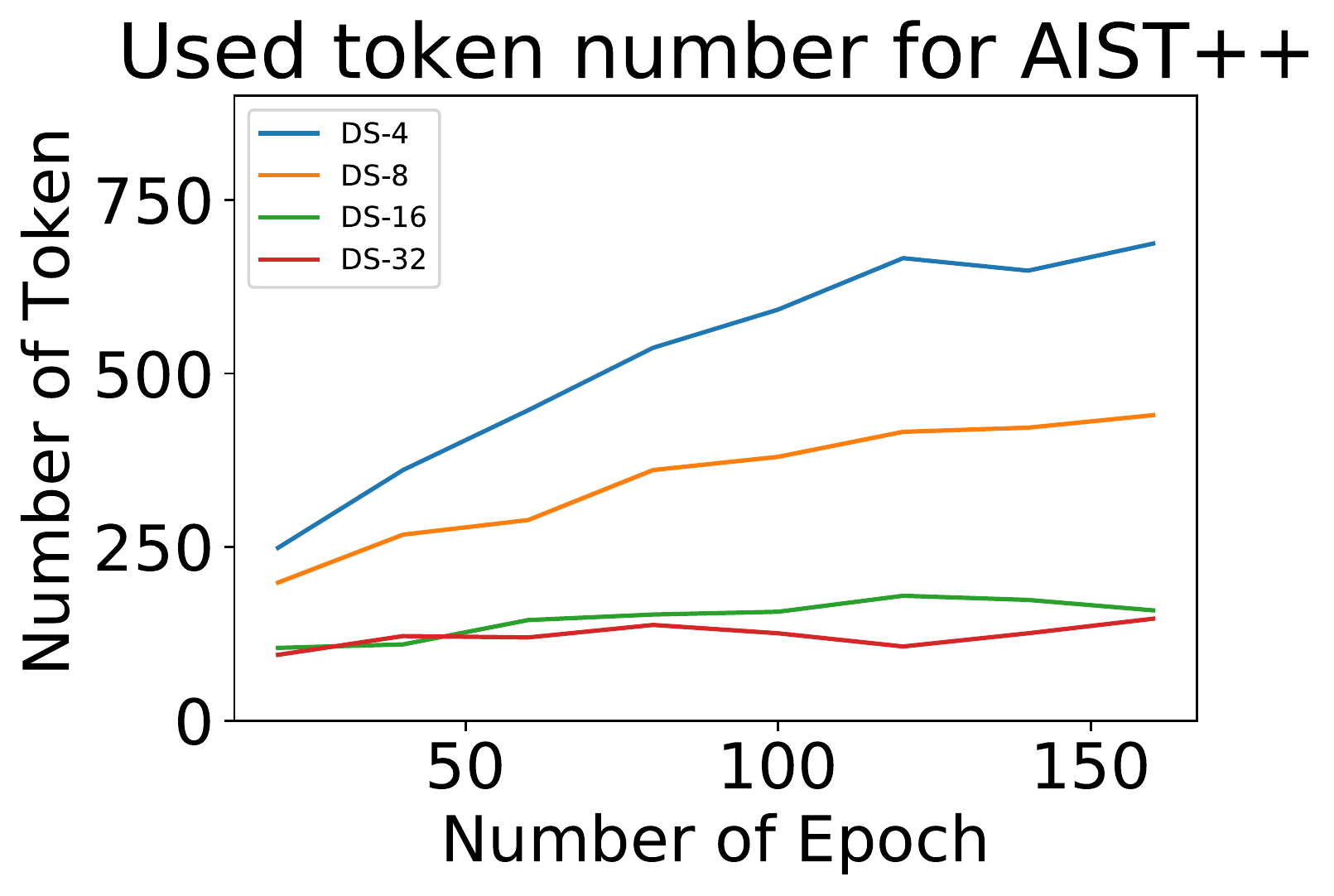}
    \end{minipage}}
    \subfloat[token-of-ml3d]{
    \begin{minipage}{0.25488\linewidth}
        \centering
        \includegraphics[width=\linewidth]{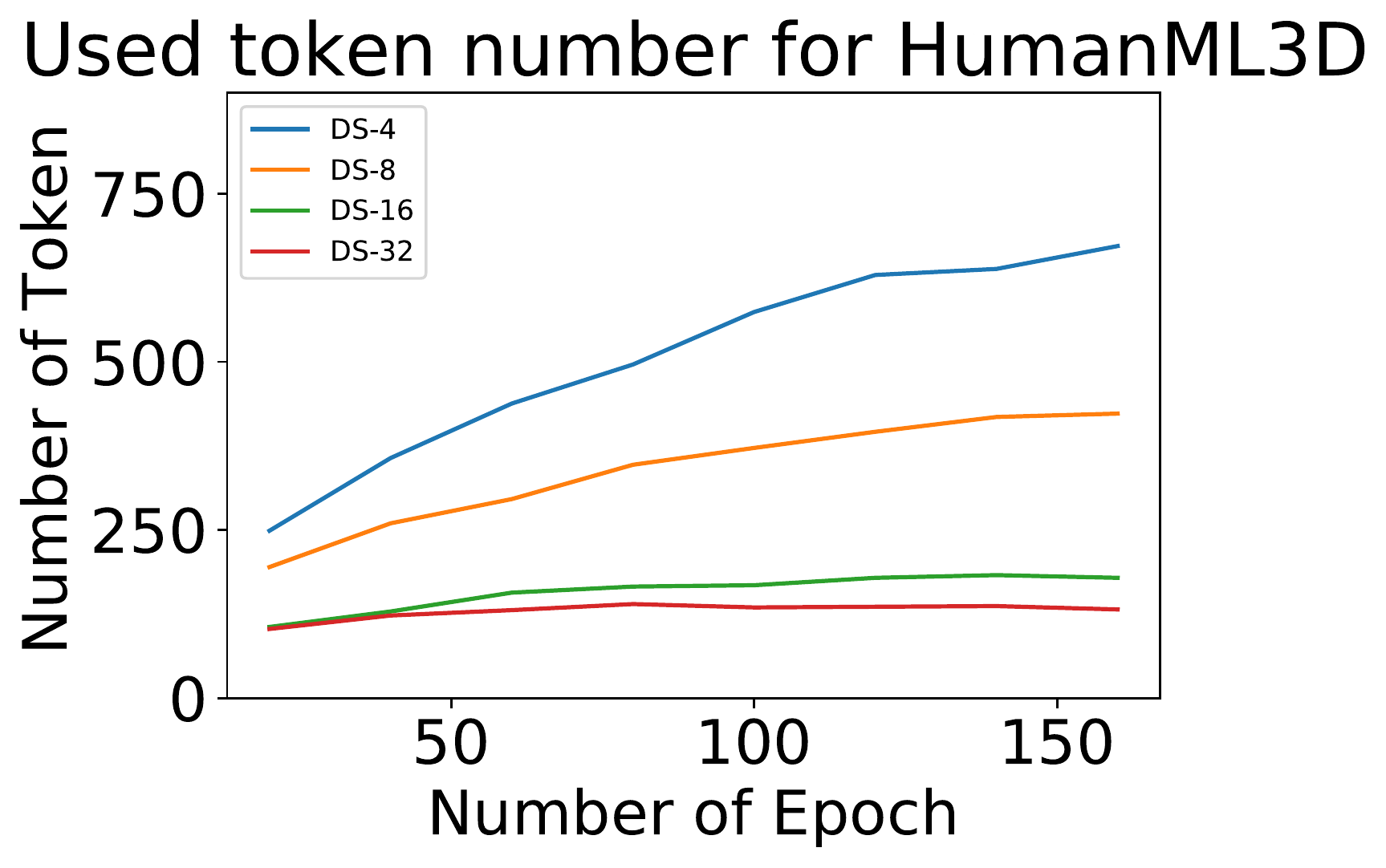}
    \end{minipage}}
    \subfloat[token-of-share]{
    \begin{minipage}{0.2346\linewidth}
        \centering
        \includegraphics[width=\linewidth]{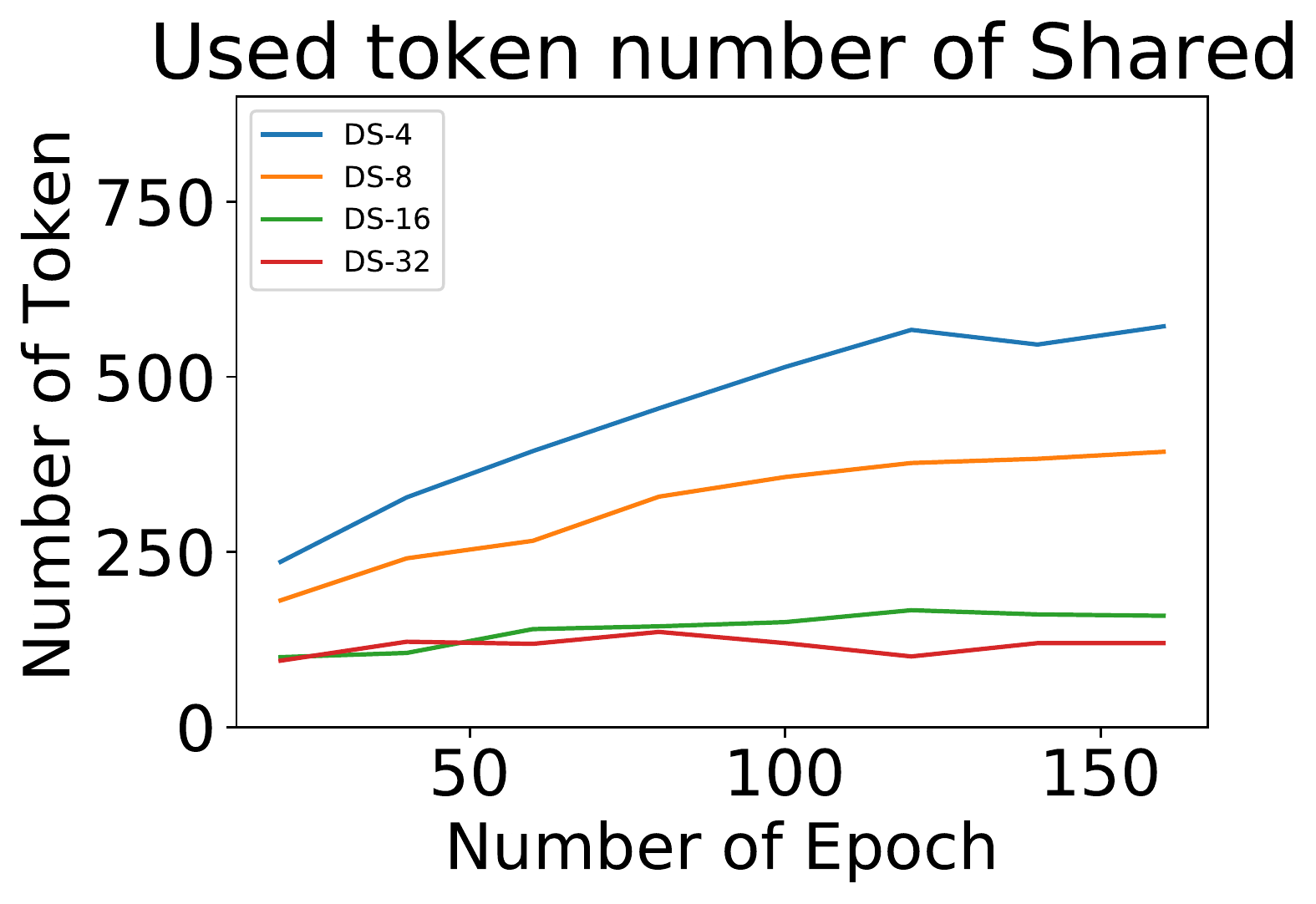}
    \end{minipage}}
    \subfloat[val-loss]{
    \begin{minipage}{0.21828\linewidth}
        \centering
        \includegraphics[width=\linewidth]{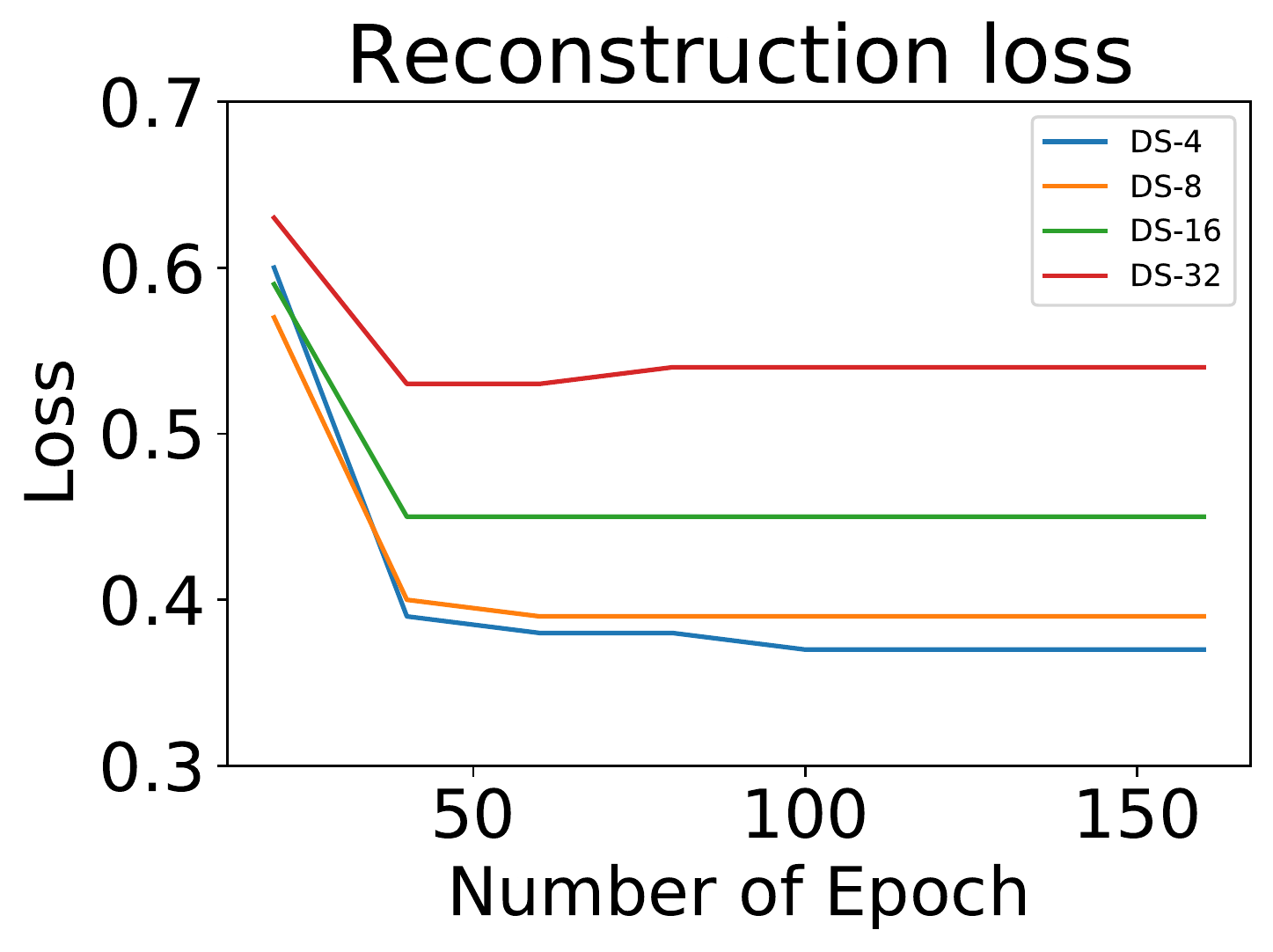}
    \end{minipage}}
    \caption{Shared tokens (latent space) with a human motion VQ-VAE architecture in Experiment B.}
    \label{fig:motion tokens}
\end{figure*}

\section{The Impact of Mixed Data}
As mentioned in the main text, a direct combination of the music2dance (AIST++ \cite{li2021ai}) and text2motion (HumanML3D \cite{guo2022generating}) in the motion space might be sub-optimal for training because the motions from these two datasets fall in completely different spaces. In contrast, we project the motions into a consistent and shared latent space with a human motion VQ-VAE architecture. To show the effectiveness of the proposed method quantitatively, we design two experiments as follows. 
\begin{itemize}
    \item Experiment A: we random sample 100 batches of data (same size as AIST++ test set) from both datasets, and measure the FID between the random batch and the whole dance data.
    \item Experiment B: we sample 30\% of the original data from both datasets and train them with a human motion VQ-VAE of different downsample rates (4, 8, 16, 32).
\end{itemize}
 
In experiment A, Figure~\ref{fig:fid-diff} shows the distribution of FID results from both datasets.
From Figure~\ref{fig:fid-diff}, we can observe that there is a distinct difference between the two datasets on geometric feature, and a small overlap in kinetic feature.

\begin{figure}[t]
	\centering
    	\includegraphics[width=\linewidth]{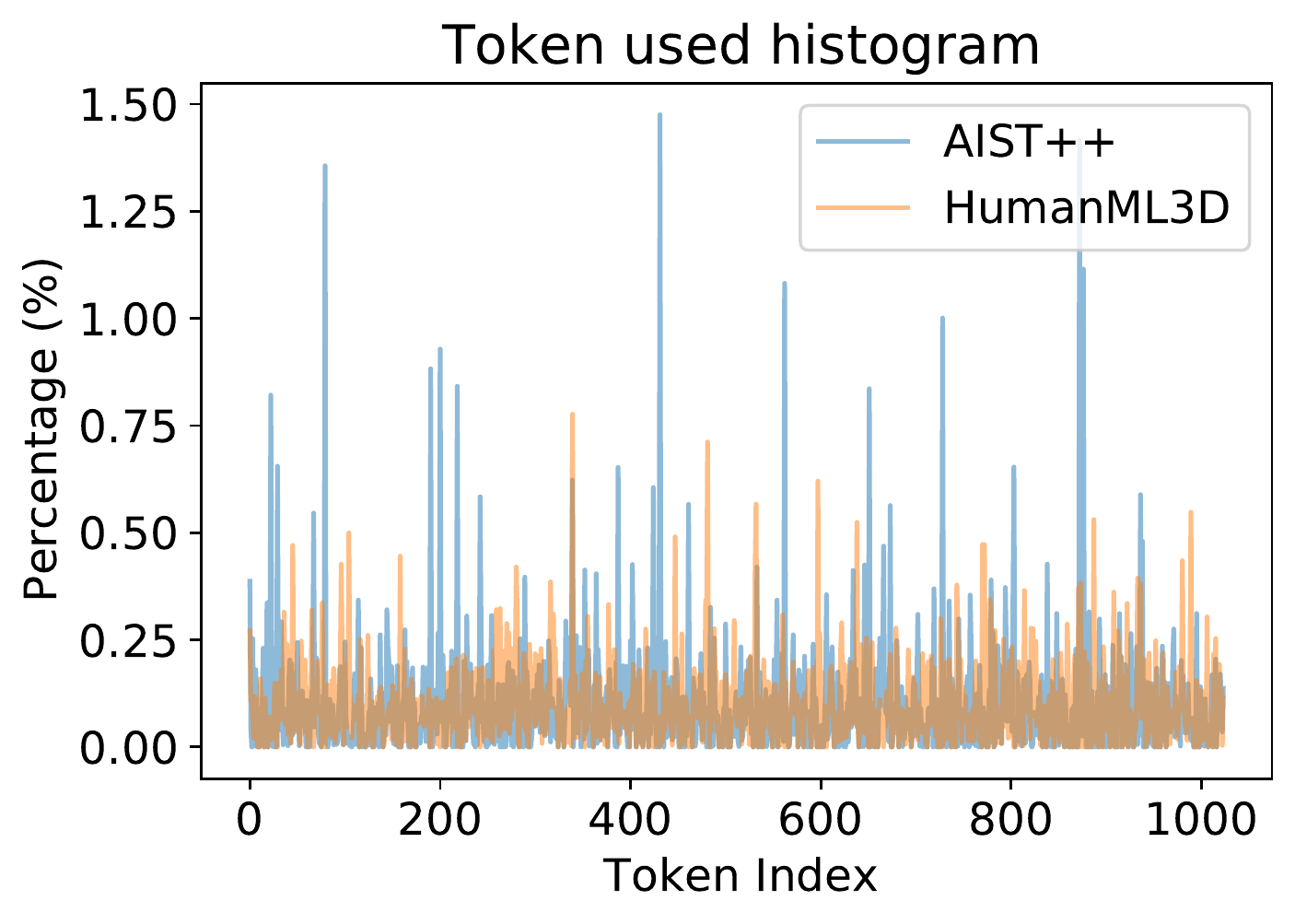}
	\caption{Token used histogram, histogram are normalized by the total frame from each dataset.}
	\label{fig:token-hist}
\end{figure}

\begin{figure}[t]
\centering
    \begin{minipage}[b]{0.5\textwidth}
        \includegraphics[width=0.95\linewidth]{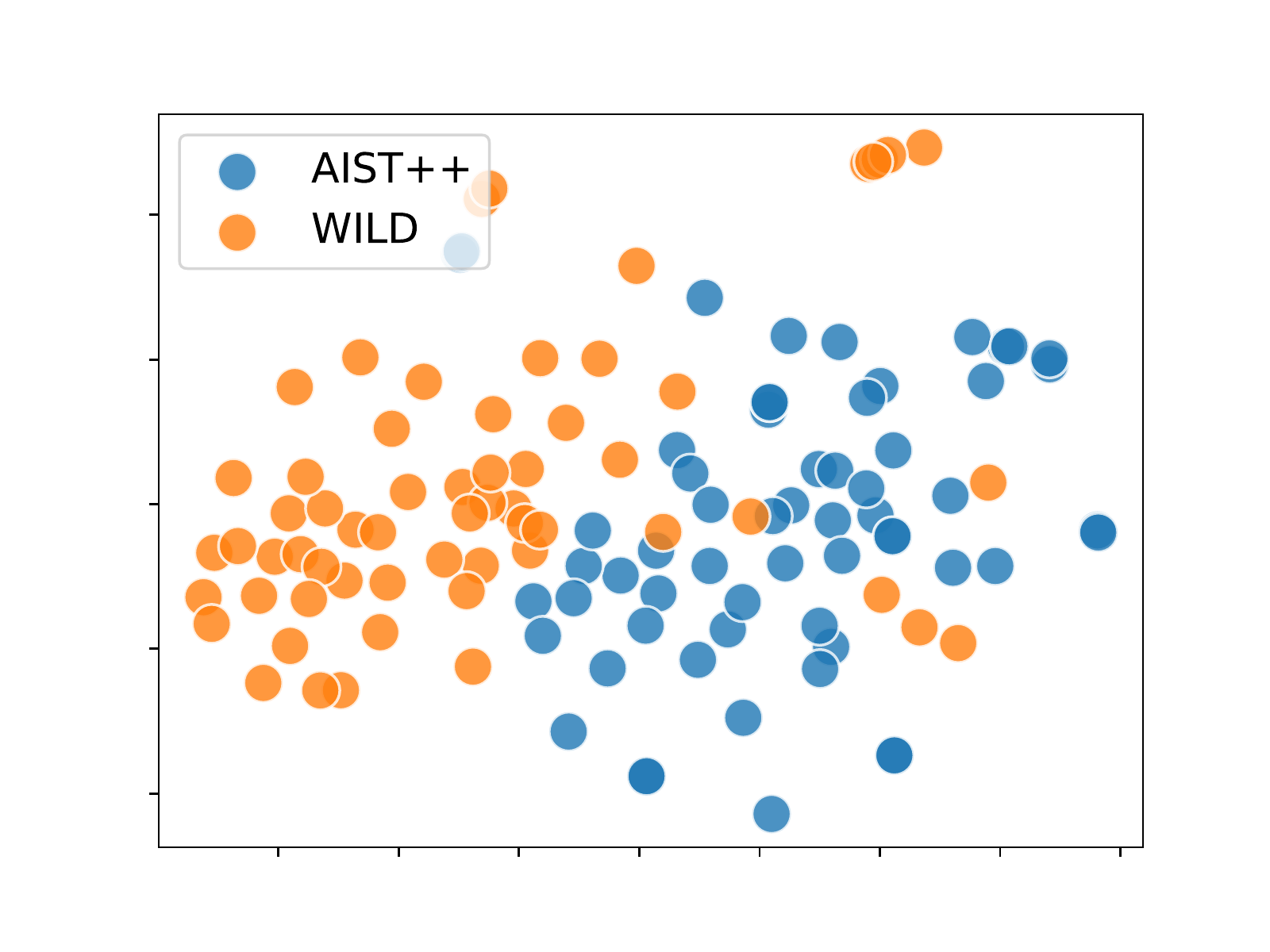}
    \end{minipage}
\caption{audio-t-SNE of datasets (orange: AIST++, blue: our dataset)). 
}
\label{fig:tsne-audio}
\end{figure}

In experiment B, from the Figure~\ref{fig:motion tokens}, we have the following three findings:
i) Figure \ref{fig:motion tokens} (a) and Figure \ref{fig:motion tokens} (b) show that the tokens used of each dataset will be increasing with the training epoch.
ii) In Figure \ref{fig:motion tokens} (c), the shared token number is also increasing together with it from both dataset.
iii) The lower the downsample rate, the higher the used token number and shared token number, with smaller reconstruction loss (val loss).
Consider that the lower the downsample rate, the longer the tokenized sequence for transformer in the second step of our pipeline. We choose downsample rate of 8, (a relatively small val loss, rich shared token number, and relatively short tokenized sequence length).

From Figure~\ref{fig:token-hist}, we can see that both datasets almost share one codebook when motions are encoded with a VQ-VAE. Specifically, the total number of vectors contained in the codebook is 1024, 855 vectors and 912 vectors of which are used to construct the motions in AIST++ and HumanML3D, respectively. 846 vectors (98.9\% in AIST++ and 92.8\% in HumanML3D) are shared to generate the motion tokens, which is much better than the feature distance from Figure~\ref{fig:fid-diff}.

\section{The Analysis of the Collected Dataset}
To verify the domain gap between source music and wild music, 
We sample the music features extracted by the Librosa (used in framework training) 
and plot a t-SNE in Figure~\ref{fig:tsne-audio}. Two music datasets lay on two different distributions with a few overlaps, which shows the generalization ability of our method.
The inferior results in 
Table~\textcolor{red}{1} (main text)
compared with our mix training show that mix gains better generalization performance.

\section{The Fusion Weight and Text2motion Results}
\begin{figure}[h]
    \centering
    \subfloat[top-1]{
    \begin{minipage}{0.5\linewidth}
        \centering
        \includegraphics[width=\linewidth]{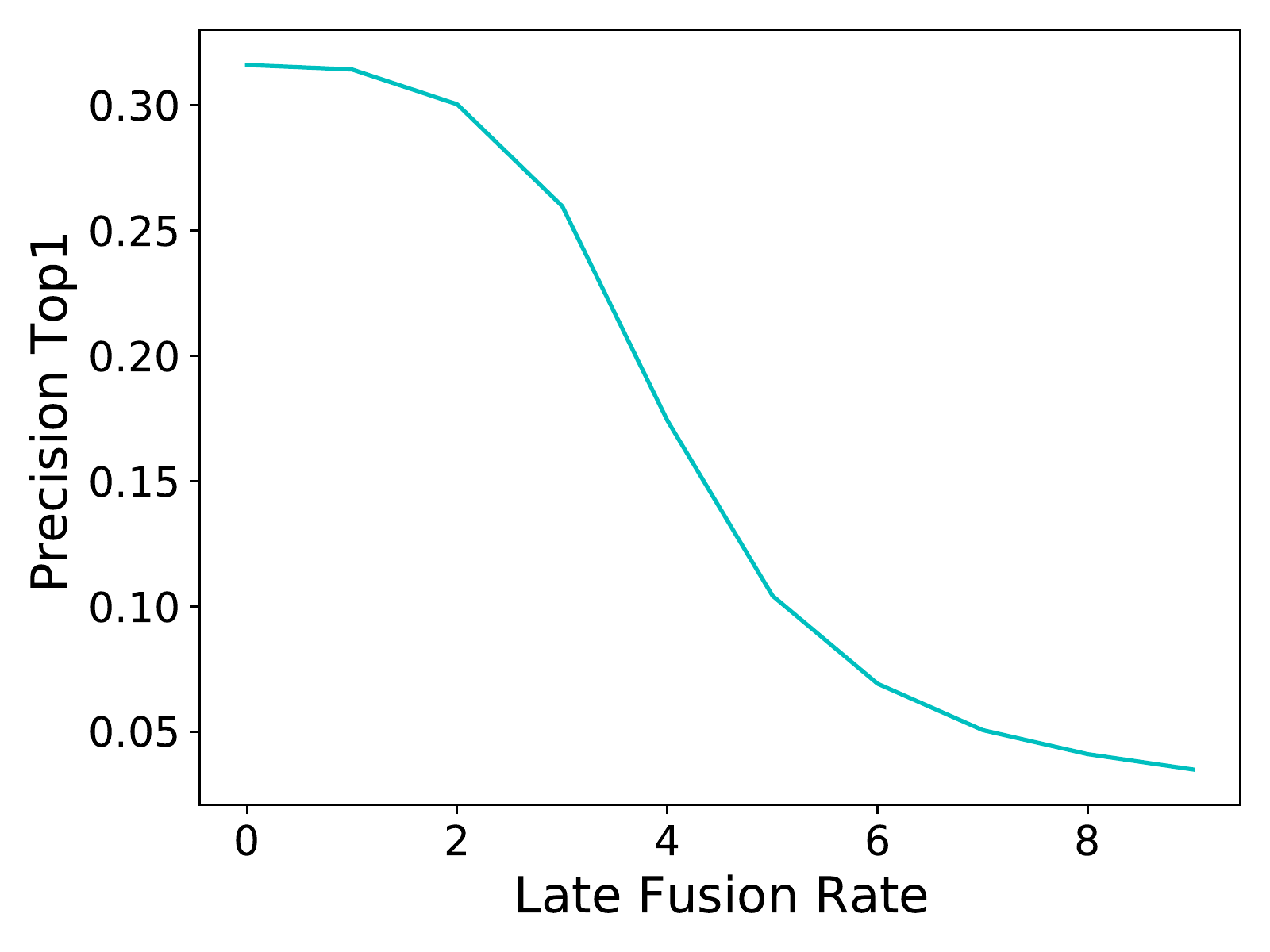}
    \end{minipage}}
    \subfloat[MM dist]{
    \begin{minipage}{0.5\linewidth}
        \centering
        \includegraphics[width=\linewidth]{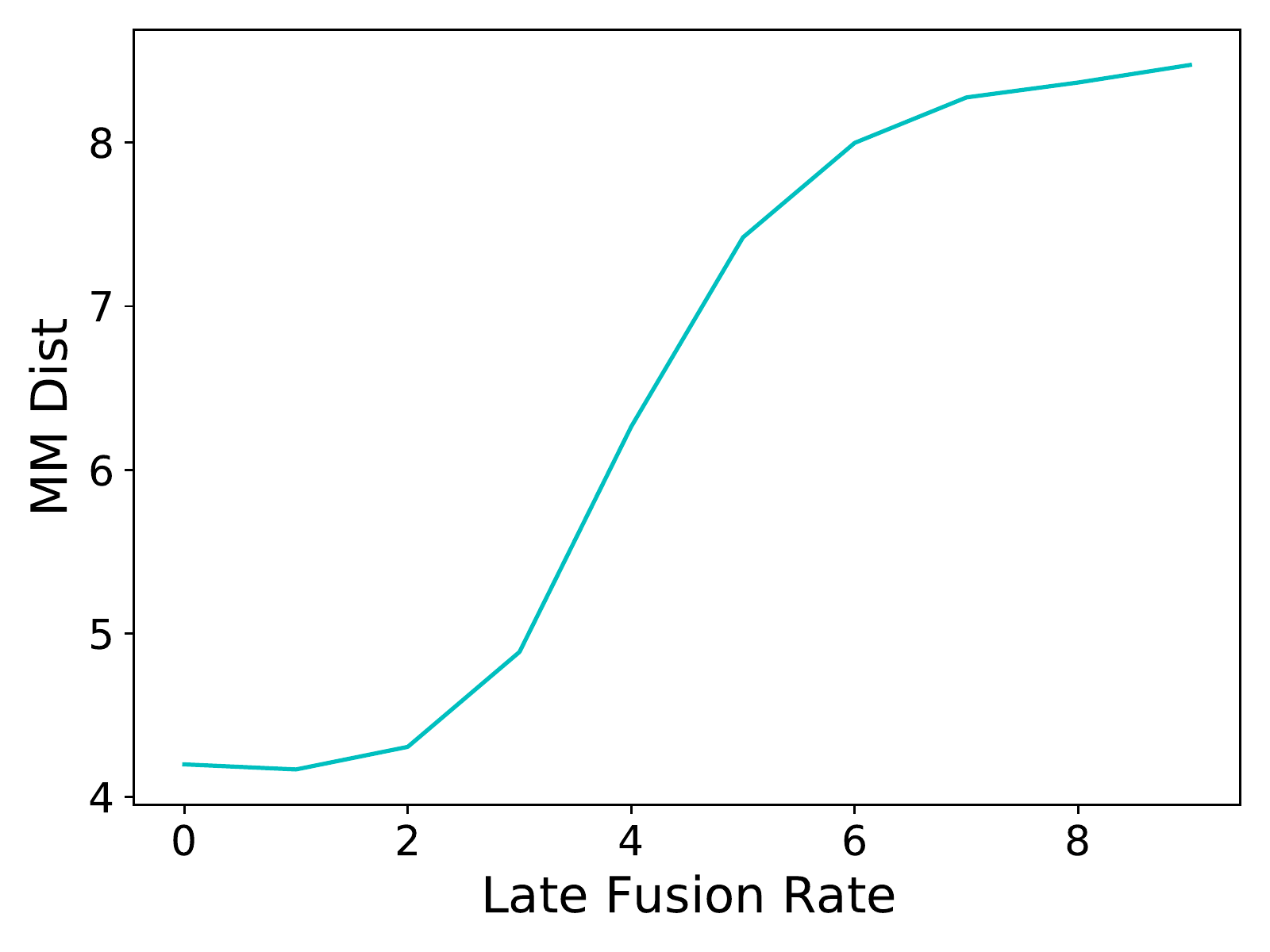}
    \end{minipage}}
    \caption{The effect of LFR with t2m resulst. 
    }
    \label{fig:lfr}
\end{figure}

We further explore the effect of late fusion rate (LFR), as shown in Fig~\ref{fig:lfr}, with the increasing of LFR, the MM distance and Top 1 precision get worse. To balance the feature content, we choose late fusion rate of 0.8.

\begin{table}[h]
\centering
{
\begin{tabular}{lllll}
\toprule
train & \multicolumn{2}{l}{AIST++} & \multicolumn{2}{l}{mix data} \\
\midrule
test  & AIST++        & wild       & AIST++         & wild        \\
\midrule
1     & 4.08 / 4.08  & 3.76 / 3.40   & 5.67 / 4.88   & 1.21 / 0.87 \\
10    & 1.31 / 1.38  & 0.61 / 0.63   & 2.58 / 2.10   & 0.10 / 0.12 \\ 
100   & 0.00 / 0.00  & 0.78 / 0.75   & 0.00 / 0.00   & 0.12 / 0.11 \\
\bottomrule
\end{tabular}
}
\caption{PFF/AUC$_f$ with topk=1, 10, 100. }
\label{table:pff-by-topk}
\end{table}

\section{Analysis of the Freeze Improvement.}
Since our method gains better results in freeze issues, we hypothesize the improvement is brought by both the architecture design and mixed training method.
We report the PFF in Table~\ref{table:pff-by-topk}.
In architecture, we {sample} tokens from the top-k tokens with the highest probability, instead of choosing the one with maximum probability as Bailando~\cite{siyao2022bailando}, which reduces the PFF. 
With extra HumanML3D data, the share motion decoder learns more motion sequence statics. Thus the PFF further improved. 
Thus both architecture and extra data mix training improve the PFF (AUC same).

\section{More Visualizations of Our Results}
We also show more visualizations of our results in the attached `demo.mp4' file, which contains the following contents. 
\begin{itemize}
    \item Comparisons with other music2dance methods in AIST++ test set and our in-the-wild dataset.
    \item Our results with the same music, different actions / time / durations.
    \item Comparisons with Slerp \cite{shoemake1985animating} for music-text conditioned dance generation.
\end{itemize}

From these videos, we can find that our results outperform other methods and are more realistic.

\end{document}